\pdfoutput=1
\documentclass[11pt]{article}

\usepackage[preprint]{acl}

\usepackage{times}
\usepackage{latexsym}
\usepackage{amsmath}
\usepackage{booktabs}
\usepackage{multirow}
\usepackage{graphicx}
\usepackage{subcaption}
\usepackage{amssymb}
\usepackage{enumitem}
\usepackage{dsfont}
\usepackage{tabularx}
\usepackage{longtable}
\usepackage{array}

\usepackage[table]{xcolor}
\definecolor{hhcol}{HTML}{E3E3FB}
\definecolor{tdqcol}{HTML}{DDF0EE}
\definecolor{bestcol}{HTML}{FFF3A8}
\definecolor{fam1}{HTML}{F4F6FD}
\definecolor{fam2}{HTML}{FDF7EE}
\definecolor{fam3}{HTML}{F1F8F1}
\definecolor{fam4}{HTML}{FBF1F4}
\definecolor{upcol}{HTML}{1B7F3A}
\definecolor{downcol}{HTML}{C0392B}
\newcommand{\up}[1]{{\scriptsize\textcolor{upcol}{#1}}}
\newcommand{\down}[1]{{\scriptsize\textcolor{downcol}{#1}}}
\newcommand{\same}[1]{{\scriptsize\textcolor{gray}{#1}}}

\usepackage[T1]{fontenc}
\usepackage[utf8]{inputenc}
\usepackage{microtype}
\usepackage{inconsolata}

\title{CulSTAR: Mitigating Cultural Drift through Dynamic Cultural Memory and Stage-Aware Reinforcement}

\author{
  \textbf{Weihua Zheng\textsuperscript{1}},
  \textbf{Bryan Chen Zhengyu Tan\textsuperscript{2}},
  \textbf{Weiwen Xu\textsuperscript{3}},
  \textbf{Hao Wu\textsuperscript{4}},
  \textbf{Rui Cao\textsuperscript{2}},
\\
  \textbf{Xiyan Tao\textsuperscript{1}},
  \textbf{Xing Xie\textsuperscript{5}},
  \textbf{Xiaoyuan Yi\textsuperscript{6}},
  \textbf{Jing Yao\textsuperscript{6}},
  \textbf{Nancy F. Chen\textsuperscript{1}},
\\
  \textbf{Zhengyuan Liu\textsuperscript{1}},
  \textbf{Huadan Wu\textsuperscript{7}},
  \textbf{Thong T. Doan\textsuperscript{2}},
  \textbf{Bich Ngoc Doan\textsuperscript{8}},
  \textbf{Roy Ka-Wei Lee\textsuperscript{9}}
\\
\\
  \textsuperscript{1}Agency for Science, Technology and Research (A*STAR),
\\
  \textsuperscript{2}Singapore University of Technology and Design,
  \textsuperscript{3}Apodex,
  \textsuperscript{4}Jiangxi Normal University,
\\
  \textsuperscript{5}Shanghai AI Laboratory,
  \textsuperscript{6}Microsoft Research Asia,
  \textsuperscript{7}China University of Petroleum,
\\
  \textsuperscript{8}EPFL,
  \textsuperscript{9}University of British Columbia
\\
  \small{
    \textbf{Correspondence:}
    \href{mailto:zhengw@a-star.edu.sg}{zhengw@a-star.edu.sg},
    \href{mailto:roy.lee@ubc.ca}{roy.lee@ubc.ca}
  }
}

\begin{document}
\maketitle
\begin{abstract}
Large language models (LLMs) may weaken or contradict previously established cultural constraints during multi-turn interactions, a phenomenon we call cultural drift. We introduce CulSTAR, which combines Dynamic Cultural Memory with stage-aware reinforcement learning to supervise memory updates, evidence acquisition, reflection, and revision. We also introduce CultureConverse-Long, a multilingual benchmark with conversations of up to 20 turns, and an anchor-based evaluation of constraint maintenance and violation. Despite training only on short conversations with reference memory, CulSTAR improves response quality and cultural consistency over base models across three LLM backbones in English and native-language interactions. Gains reach 26.4 percentage points in anchor maintenance and 31.2 percentage points in survival through five applicable turns, with lower contradiction rates.
\end{abstract}

\section{Introduction}

Culturally appropriate assistance requires LLMs to preserve and apply cultural constraints across conversational turns. Existing benchmarks reveal persistent gaps in cultural knowledge, adaptability, and grounded reasoning across social contexts~\citep{shi2024culturebank,feng2025culfit}. In multi-turn conversations, individually appropriate responses may violate earlier cultural constraints. We refer to the weakening, replacement, or contradiction of previously established cultural constraints while they remain applicable as \emph{cultural drift}.

Preventing \emph{cultural drift} presents several challenges. First, cultural information emerges gradually and implicitly, obscuring which identities, preferences, and constraints should persist. Second, cultural constraints are temporally sparse and may become decision-critical many turns later, while retention alone does not ensure correct application. Third, cultural state is non-monotonic as new information refines, supersedes, or invalidates earlier assumptions. Prior work has documented drift from established personas and behavioral constraints in extended LLM conversations~\citep{li2024measuring,choi2024examining}. Unlike static personas, cultural grounding requires assistants to continually update and apply cultural constraints as they emerge and evolve across dialogue turns. However, existing cultural alignment approaches primarily emphasize response-level appropriateness and norm conformity, providing limited supervision for intermediate decisions such as updating cultural beliefs, resolving conflicting evidence, and preserving relevant constraints across turns~\citep{luo2025agent,tao2026trace}. Without such supervision, a response-level signal cannot indicate which intermediate decision went wrong, and a response that appears appropriate may rest on an incorrect cultural state whose consequences only surface turns later.

To address this challenge, we introduce \textbf{CulSTAR}, a culturally grounded agentic learning framework that treats cultural interaction as a sequential state-maintenance problem. Its Dynamic Cultural Memory tracks evolving cultural context, while Stage-Aware Cultural Credit Assignment with GRPO~\citep{shao2024deepseekmath} (SCCA-GRPO) provides stage-specific learning signals for memory transitions, culturally grounded evidence acquisition, reflection, and revision. These components extend cultural supervision beyond the final response to the intermediate decisions that support sustained cultural consistency. SCCA-GRPO decomposes each turn into stages, with stage-level training conditioned on a reference pre-turn memory, thereby reducing training cost without requiring full-conversation optimization. Long-horizon consistency then emerges from the composition of correct local state transitions through the explicit memory.
To study cultural drift over extended interactions, we construct \textit{CultureConverse-Long}, a multilingual dataset of conversations extending to 20 turns, with scenario specifications covering cultural norms, taboos, and cross-turn requirements. We further introduce an anchor-based longitudinal evaluation framework that maps constraints to applicable turns and tracks maintenance, first violation, survival, and recovery. This evaluates whether established cultural constraints remain behaviorally effective throughout a conversation, beyond the appropriateness of individual responses.

Across three LLM backbones and both language settings, CulSTAR improves conversational quality over base models and supervised adaptation. On long conversations, it also strengthens cultural anchor maintenance and survival while reducing cross-turn contradictions, demonstrating more durable culturally grounded behavior, even though training uses only short conversations.

Our contributions are threefold:

\begin{itemize}[leftmargin=*,topsep=0pt] \setlength{\itemsep}{0pt} \setlength{\parskip}{0pt} \setlength{\parsep}{0pt}

\item \textbf{CulSTAR framework.} We combine explicit cultural state maintenance with stage-specific supervision of culturally grounded agentic interaction.

\item \textbf{CultureConverse-Long dataset.} We construct a multilingual long-conversation dataset with explicit cultural constraints and cross-turn requirements, providing a testbed for studying cultural drift across models and language settings.

\item \textbf{Longitudinal evaluation.} We introduce an anchor-based framework to assess cultural drift across turns by tracking cultural constraint maintenance, violation, survival, and recovery.

\end{itemize} \noindent

\section{Related Work}

\paragraph{Cultural competence and interactive evaluation.}

Existing benchmarks reveal limitations in LLM cultural knowledge, adaptability, and reasoning across social contexts\citep{chiu2025culturalbench,rao2025normad,shi2024culturebank}. Recent evaluations extend this focus to multilingual, multi-turn interactions\citep{wu2025socialcc,tan2026cultureconverse,pham2026liveculturebench}. In this work, we examine whether established cultural constraints remain behaviorally effective throughout a conversation.

\paragraph{Long-term consistency and cultural memory.}

Related studies show that LLMs can gradually deviate from previously established personas or behavioral constraints during extended conversations \citep{li2024measuring,choi2024examining}. Memory-augmented systems and benchmarks study information persistence, updating, and temporal reasoning over dialogue histories \citep{zhong2024memorybank,wu2024longmemeval}. Our Cultural Memory similarly maintains evolving conversational state, but explicitly tracks the validity and evolution of culturally relevant information. Cultural drift, however, is not equivalent to forgetting: retained cultural facts may no longer influence later decisions. We therefore examine whether cultural state remains behaviorally operative throughout the interaction.

\paragraph{Agentic reinforcement learning and cultural credit assignment.}

Long-horizon LLM agents increasingly rely on reinforcement learning over trajectories involving reasoning, retrieval, and tool use, making intermediate credit assignment a central challenge. Agent Lightning and TRACE provide localized trajectory supervision \citep{luo2025agent,tao2026trace}, while CulFiT introduces fine-grained rewards for culturally grounded response alignment \citep{feng2025culfit}. These directions address process supervision and cultural alignment largely separately. Our Stage-Aware Cultural Credit Assignment connects them through evolving cultural state, supervising memory transitions, tool use, and self-correction at each stage of a turn.

\section{CulSTAR framework}

\begin{figure*}[t]
  \centering
  \includegraphics[width=\textwidth]{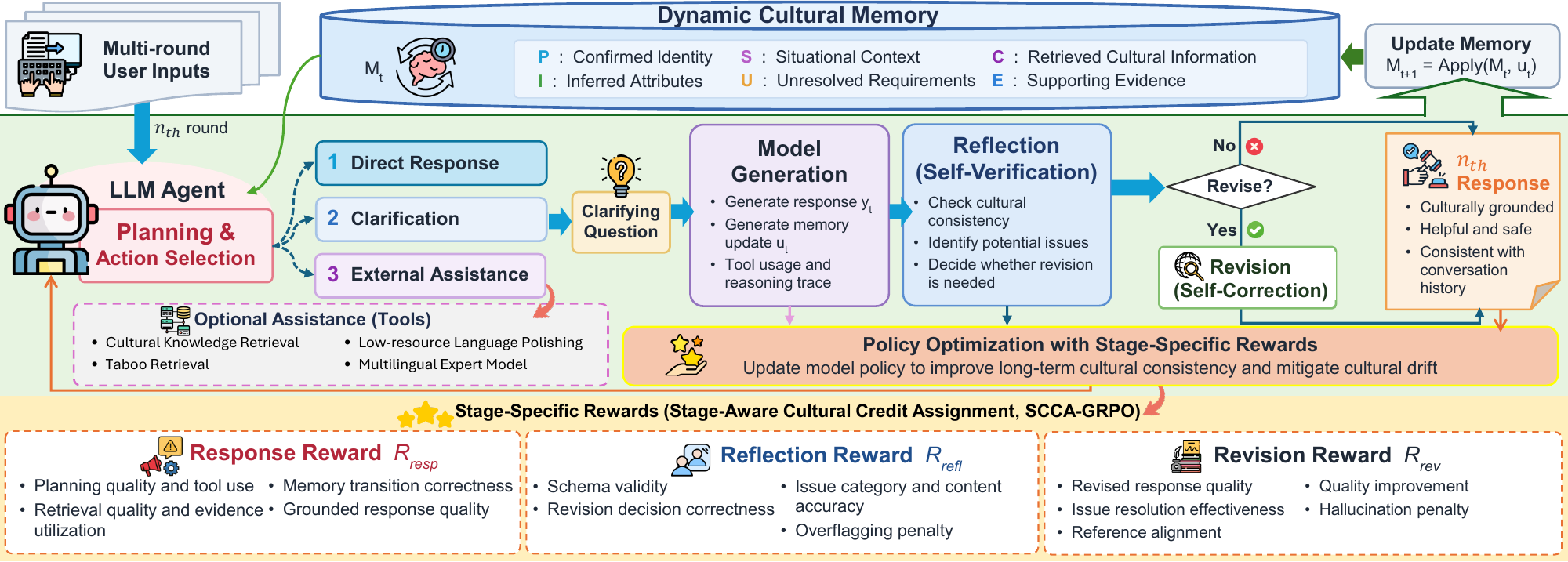}
  \captionsetup{skip=2pt} 
  \caption{Overview of CulSTAR with Dynamic Cultural Memory and SCCA-GRPO. The figure illustrates the workflow of a single turn within a multi-turn conversation.}
  \label{fig:overview}
\end{figure*}

\subsection{Problem Formulation}

We formulate culturally grounded multi-turn interaction as a sequential decision problem centered on cultural state maintenance. At turn $t$, the agent receives the user input, conversation history, and Cultural Memory $M_t$. It must preserve established cultural information, update it in light of new evidence, and retrieve external knowledge when needed to generate contextually consistent responses. Our objective is to jointly optimize the final response and the intermediate decisions supporting cultural state maintenance, evidence acquisition, and inconsistency correction. Each turn produces a trajectory $\tau_t = (\text{route}, \text{tools}, \text{queries}, \text{evidence}, y_t, u_t, \text{reflection}, \tilde{y}_t)$. Errors at different stages of $\tau_t$ arise from different causes and require different corrections, yet a single response-level reward cannot distinguish them. Optimizing long conversations end to end would require costly multi-turn rollouts and would spread the credit for each decision over many turns. We therefore split learning into stages: each training turn is conditioned on a reference pre-turn memory $M_t$, which isolates the quality of current-turn decisions from errors accumulated in earlier turns, and each stage receives its own reward. Long-horizon consistency is not optimized directly. Because cultural state is carried explicitly by the memory, it can emerge from composing correct local state transitions across turns. At test time, we evaluate this with fully self-generated memory over conversations of up to 20 turns.

\subsection{Framework Overview}

CulSTAR combines Dynamic Cultural Memory to track evolving cultural context with SCCA-GRPO to supervise decisions grounded in that state.
As shown in Figure~\ref{fig:overview}, at each turn, the agent chooses direct answering, clarification, or external assistance. Using the memory and retrieved evidence, it generates a candidate response and a memory update. An independent reflection checks cultural consistency and interaction requirements, triggering revision when needed. The memory update then produces the next turn's cultural state.

\subsubsection{Dynamic Cultural Memory}

As users introduce identities, refine preferences, impose constraints, or invalidate earlier assumptions, cultural context evolves. We represent this context as an explicit, dynamically updated state. At turn $t$, Cultural Memory $M_t$ contains six slots: confirmed profiles, inferred cultural attributes with confidence estimates, situational constraints, unresolved requirements, supporting evidence, and retrieved cultural information. These slots distinguish established information, uncertain inferences, contextual requirements, and external evidence.
Each memory update $u_t$ specifies an operation, target slot, attribute-value pair, confidence estimate, and supporting evidence. The supported operations add, refine, reweight, resolve, and deprecate respectively introduce information, modify existing entries, adjust confidence, resolve requirements, and invalidate obsolete entries. Applying $u_t$ to $M_t$ produces $M_{t+1}$.

\paragraph{Turn-level memory supervision.}
To isolate current-turn update quality from errors accumulated in earlier turns, the memory reward compares changes induced by predicted and gold updates from the same pre-turn state $M_t$:
\begin{equation}
\begin{aligned}
\Delta M_t^{\mathrm{pred}} &= \operatorname{Apply}(M_t, u_t^{\mathrm{pred}}) - M_t, \\
\Delta M_t^{\mathrm{gold}} &= \operatorname{Apply}(M_t, u_t^{\mathrm{gold}}) - M_t.
\end{aligned}
\end{equation}

\subsubsection{Culturally Grounded Agentic Interaction}
\paragraph{Planning and external assistance.}
At each turn, the agent selects one of three actions: answering directly, asking a clarification question, or invoking external assistance. For the latter, we expose four assistance tools that the model may invoke through tool calls. Two target the language competence: the \emph{multilingual expert model} interprets user input when comprehension of a low-resource language is unreliable, and \emph{low-resource language polishing} improves the fluency of generated responses. The other two target cultural knowledge acquisition: \emph{cultural knowledge retrieval} supplies relevant norms and practices, and \emph{taboo retrieval} flags potentially inappropriate actions or recommendations. The returned observations condition response generation and provide learning signals for tool selection and query formulation.

\paragraph{Grounded generation.}
Given the user input, dialogue history, cultural memory, and retrieved evidence, the model generates a candidate response $y_t$ and a memory update $u_t$. The response addresses the current request, while $u_t$ captures newly established cultural information for subsequent turns.

\paragraph{Reflection and conditional revision.}
The model performs self-reflection on $y_t$ to assess cross-turn consistency, cultural and taboo compliance, requirement coverage, and evidential support. Based on assessment, if revision is needed, it produces a revised response $\tilde{y}_t$ that addresses the issues. Otherwise, $y_t$ is returned unchanged. Revision modifies only the user-visible response and leaves the memory update $u_t$ intact, keeping response correction separate from persistent memory updates.

\subsection{Stage-Aware Cultural Credit Assignment with GRPO}

A brief supervised fine-tuning warm-up on single-turn examples first teaches the output schema, including routing decisions, tool-call syntax, memory-update fields, and the reflection format. We then use SCCA-GRPO to distinguish trajectories that may produce similar final responses but differ in how they maintain state, acquire evidence, and correct errors. Its design follows three principles. \textbf{(i) Stage separation.} We construct 3,166 training instances drawn from short conversations in the CultureConverse training set~\citep{tan2026cultureconverse}, each conditioned on the reference pre-turn memory so that per-turn credit is not confounded by errors accumulated in earlier turns. For each instance, we sample $G$ end-to-end rollouts comprising response generation, reflection on $y_t$, and revision to $\tilde{y}_t$ when deemed necessary. Each rollout is partitioned into three stage-specific segments (Figure~\ref{fig:overview}), with rewards and group-relative advantages computed separately for each stage across the $G$ rollouts. Tokens are updated only with their segment's advantage, giving error detection and correction direct signals instead of folding them into one outcome score.
\textbf{(ii) Hybrid credit.} Structured gold annotations support deterministic scoring, while properties requiring semantic judgment are evaluated by an LLM judge. This keeps the process credit largely verifiable and limits dependence on the judge.
\textbf{(iii) Explicit weighting.} Within each stage, reward components are normalized to the unit interval, combined with fixed coefficients, and logged separately to expose behavioral regressions and support weight recalibration. We detail each stage's reward components below.

\paragraph{Response reward.}
The response stage covers planning, evidence acquisition, grounded generation, and memory updating:
\begin{equation} R_{\mathrm{resp}} = R_{\mathrm{rule}} + R_{\mathrm{judge}} - P_{\mathrm{resp}}. \label{eq:r_resp} \end{equation}

$R_{\mathrm{rule}}$ supervises planning, tool selection, retrieval queries, clarification, language assistance, and memory transitions. Retrieval quality considers cultural entity coverage and topical relevance, while memory correctness is assessed through the state transition $\Delta M_t$ from the pre-turn memory $M_t$, including updated attributes, fields, and values. $R_{\mathrm{judge}}$ evaluates goal completion, cultural consistency, taboo avoidance, clarification quality, language adequacy, and memory-update correctness. $P_{\mathrm{resp}}$ penalizes unnecessary tool calls, duplicate memory additions, unsupported claims, and regressions in critical cultural behaviors.

\paragraph{Reflection reward.}
Reflection is optimized independently so that error detection receives direct supervision:
\begin{equation} \begin{aligned} R_{\mathrm{refl}} ={}& s_{\mathrm{schema}} \bigl( \alpha_1 R_{\mathrm{required}} +\alpha_2 R_{\mathrm{category}} \\ &\qquad +\alpha_3 R_{\mathrm{content}} \bigr) -P_{\mathrm{overflag}} . \end{aligned} \label{eq:r_refl} \end{equation}

Here, $s_{\mathrm{schema}}$ gates the reward by structural validity. $R_{\mathrm{required}}$ evaluates whether revision is correctly triggered, $R_{\mathrm{category}}$ whether the appropriate category is identified, and $R_{\mathrm{content}}$ whether the reported issue is genuine and substantive. $P_{\mathrm{overflag}}$ discourages unsupported issue detection.

\paragraph{Revision reward.}
The revision stage receives both the original response and its reflection, and is trained to repair identified issues while preserving already correct content:
\begin{equation} \begin{aligned} R_{\mathrm{rev}} ={}& \beta_1 Q_{\mathrm{new}} +\beta_2 R_{\mathrm{resolve}} +\beta_3 R_{\mathrm{align}} \\ &+ \beta_4 R_{\Delta} -\beta_5 P_{\mathrm{hall}} . \end{aligned} \label{eq:r_rev} \end{equation}

$Q_{\mathrm{new}}$ measures revised-response quality, $R_{\mathrm{resolve}}$ whether identified issues are repaired, and $R_{\mathrm{align}}$ semantic consistency with the reference revision. $R_{\Delta}$ captures the quality change relative to the original response, assigning stronger penalties to regressions than rewards to improvements of equal magnitude. $P_{\mathrm{hall}}$ penalizes unsupported information introduced during revision. Detailed reward coefficients, penalty terms, and the construction of $R_{\Delta}$ are provided in App.~\ref{app:reward_details}.

These stage-specific rewards assign cultural credit to state maintenance, evidence acquisition, error detection, and correction, extending supervision beyond a single terminal response-level signal.

\section{CultureConverse-Long Dataset and Longitudinal Evaluation}
\label{sec:cultural_drift}

\paragraph{CultureConverse-Long Dataset.} 

Following the CultureConverse~\citep{tan2026cultureconverse} construction protocol, we develop CultureConverse-Long with 500 culturally grounded scenarios evaluated in English and native languages, yielding 1,000 test instances across 10 regions and 10 languages. Conversations extend up to 20 turns, guided by blueprints specifying cultural norms, taboos, and cross-turn constraints. The dataset contains 3,090 cultural anchors, averaging 6.18 per scenario. Of these, 60.0\% recur across interaction shards compared with 50.4\% in the shorter setting, with an average span of 7.72 shards between their first and final designed occurrences. We align these anchors with dialogue turns to evaluate cultural maintenance, contradictions, and first violations. App.~\ref{app:Dataset construction} provides construction details and statistics.

\paragraph{Evaluation framework.}
 We evaluate \textit{cultural drift} at the level of anchor--turn pairs, where each cultural anchor corresponds to a self-contained and verifiable requirement derived from the scenario specification, such as a taboo, ritual norm or context-specific behavioral constraint. The same cached anchors and anchor-to-turn mappings are shared across all systems.
Evaluation proceeds in two stages. First, for each anchor $a$ and turn $t$, we determine whether the anchor is applicable to the current turn. Explicit dataset mappings are used when available. Otherwise, an LLM judge performs a response-blind applicability assessment using the anchor, supporting evidence, the current user message, and only the dialogue history needed to resolve prior commitments. The assistant response is excluded at this stage to prevent applicability from being influenced by model behavior.

For applicable pairs, a second judge determines whether the response maintains or violates the anchor. Explicitly applying an anchor or remaining implicitly consistent with it is treated as maintenance, whereas attenuation, replacement by a different cultural framework, or direct contradiction is treated as violation. We use GPT-5.5 for both LLM-based evaluation stages, with judge comparisons and human validation detailed in App.~\ref{app:human evaluation}. Let $M_{a,t}\in\{0,1\}$ denote this binary maintenance outcome. Our primary metric is the micro-averaged \textit{Anchor Maintenance Rate (AMR)},
\begin{equation}
    \mathrm{AMR}_{\mathrm{micro}}
    =
    \frac{
        \sum_{(a,t)\in\mathcal{P}} M_{a,t}
    }{
        |\mathcal{P}|
    },
\end{equation}
where $\mathcal{P}$ contains all applicable anchor--turn pairs.

To capture temporal drift, we track each anchor across applicable turns and measure its first violation. Kaplan--Meier analysis estimates cultural-consistency survival, reporting the median survival turn and $\hat{S}(3)$ and $\hat{S}(5)$, which measure violation-free survival through three and five applicable turns. AMR captures average maintenance, while survival measures durability before the first violation. Finally, to distinguish temporary violations from sustained cultural drift, we further assess recovery rate and maximum violation-run length. Recovery rate measures whether a violated anchor returns to a maintained state and remains so at its final applicable turn, while maximum violation-run length captures the longest consecutive sequence of violations. App.~\ref{app:cultural_drift} provides further details on anchor definitions, annotations, additional metrics, violation decomposition, and uncertainty estimation.

\begin{table}[t]
\centering
\scriptsize
\setlength{\tabcolsep}{1.5pt}
\renewcommand{\arraystretch}{1.0}
\begin{tabular}{
    l
    @{\hspace{4pt}}
    l
    @{\hspace{3pt}}
    c
    @{\hspace{3pt}}
    c
    @{\hspace{1pt}}
    c
}
\toprule
\rowcolor{gray!12}
\textbf{Backbone} & \textbf{Method}
& \cellcolor{hhcol}\textbf{3H}
& \cellcolor{tdqcol}\textbf{TDQ}
& \textbf{Clean Rate (\%)} \\
\midrule

\multicolumn{5}{c}{\cellcolor{gray!12}\textit{English}} \\
\midrule

\rowcolor{fam2}
& Base
& \cellcolor{hhcol}3.35
& \cellcolor{tdqcol}3.49
& 67.2 \\
\rowcolor{fam2}
& + SFT
& \cellcolor{hhcol}3.36\kern1pt \up{+0.3\%}
& \cellcolor{tdqcol}3.45\kern1pt \down{-1.1\%}
& 75.8\kern1pt \up{+12.8\%} \\
\rowcolor{fam2}
\multirow{-3}{*}{Qwen3-8B-IT}
& CulSTAR
& \cellcolor{bestcol}\textbf{3.49}\kern1pt \up{+4.2\%}
& \cellcolor{bestcol}\textbf{3.51}\kern1pt \up{+0.6\%}
& \cellcolor{bestcol}\textbf{79.2}\kern1pt \up{+17.9\%} \\
\cmidrule(lr){1-5}

\rowcolor{fam3}
& Base
& \cellcolor{hhcol}3.03
& \cellcolor{tdqcol}3.35
& 70.2 \\
\rowcolor{fam3}
& + SFT
& \cellcolor{hhcol}3.19\kern1pt \up{+5.3\%}
& \cellcolor{tdqcol}3.35\kern1pt \same{0.0\%}
& 67.4\kern1pt \down{-4.0\%} \\
\rowcolor{fam3}
\multirow{-3}{*}{Llama-3.1-8B-IT}
& \textbf{CulSTAR}
& \cellcolor{bestcol}\textbf{3.38}\kern1pt \up{+11.6\%}
& \cellcolor{bestcol}\textbf{3.44}\kern1pt \up{+2.7\%}
& \cellcolor{bestcol}\textbf{78.4}\kern1pt \up{+11.7\%} \\
\cmidrule(lr){1-5}

\rowcolor{fam4}
& Base
& \cellcolor{hhcol}2.83
& \cellcolor{tdqcol}3.12
& 75.2 \\
\rowcolor{fam4}
& + SFT
& \cellcolor{hhcol}2.82\kern1pt \down{-0.4\%}
& \cellcolor{tdqcol}3.10\kern1pt \down{-0.6\%}
& \cellcolor{bestcol}\textbf{78.4}\kern1pt \up{+4.3\%} \\
\rowcolor{fam4}
\multirow{-3}{*}{SEA-LION-v4-8B-IT}
& CulSTAR
& \cellcolor{bestcol}\textbf{3.27}\kern1pt \up{+15.5\%}
& \cellcolor{bestcol}\textbf{3.40}\kern1pt \up{+9.0\%}
& 77.6\kern1pt \up{+3.2\%} \\

\midrule
\multicolumn{5}{c}{\cellcolor{gray!12}\textit{Native}} \\
\midrule

\rowcolor{fam2}
& Base
& \cellcolor{hhcol}3.23
& \cellcolor{tdqcol}3.41
& 77.0 \\
\rowcolor{fam2}
& + SFT
& \cellcolor{hhcol}3.38\kern1pt \up{+4.6\%}
& \cellcolor{tdqcol}3.42\kern1pt \up{+0.3\%}
& 77.4\kern1pt \up{+0.5\%} \\
\rowcolor{fam2}
\multirow{-3}{*}{Qwen3-8B-IT}
& CulSTAR
& \cellcolor{bestcol}\textbf{3.47}\kern1pt \up{+7.4\%}
& \cellcolor{bestcol}\textbf{3.45}\kern1pt \up{+1.2\%}
& \cellcolor{bestcol}\textbf{77.8}\kern1pt \up{+1.0\%} \\
\cmidrule(lr){1-5}

\rowcolor{fam3}
& Base
& \cellcolor{hhcol}2.46
& \cellcolor{tdqcol}2.98
& 63.0 \\
\rowcolor{fam3}
& + SFT
& \cellcolor{hhcol}2.73\kern1pt \up{+11.0\%}
& \cellcolor{tdqcol}3.00\kern1pt \up{+0.7\%}
& 61.6\kern1pt \down{-2.2\%} \\
\rowcolor{fam3}
\multirow{-3}{*}{Llama-3.1-8B-IT}
& CulSTAR
& \cellcolor{bestcol}\textbf{3.24}\kern1pt \up{+31.7\%}
& \cellcolor{bestcol}\textbf{3.34}\kern1pt \up{+12.1\%}
& \cellcolor{bestcol}\textbf{73.0}\kern1pt \up{+15.9\%} \\
\cmidrule(lr){1-5}

\rowcolor{fam4}
& Base
& \cellcolor{hhcol}2.76
& \cellcolor{tdqcol}3.06
& 74.2 \\
\rowcolor{fam4}
& + SFT
& \cellcolor{hhcol}2.80\kern1pt \up{+1.4\%}
& \cellcolor{tdqcol}3.04\kern1pt \down{-0.7\%}
& 73.8\kern1pt \down{-0.5\%} \\
\rowcolor{fam4}
\multirow{-3}{*}{SEA-LION-v4-8B-IT}
& CulSTAR
& \cellcolor{bestcol}\textbf{3.29}\kern1pt \up{+19.2\%}
& \cellcolor{bestcol}\textbf{3.35}\kern1pt \up{+9.5\%}
& \cellcolor{bestcol}\textbf{77.2}\kern1pt \up{+4.0\%} \\

\bottomrule
\end{tabular}
\captionsetup{skip=4pt} 
\caption{Results on the CultureConverse test set~\citep{tan2026cultureconverse}. Clean Rate is the percentage of clean episodes among 500. Small numbers show relative changes from the corresponding base model. Best values are bolded and highlighted. Full results in App.~\ref{app:detailed_results}.}
\label{tab:main_results_shorter}
\end{table}

\newcommand{\baseCI}[3]{%
  #1\kern1pt{\tiny [#2, #3]}}

\newcommand{\oursCI}[4]{%
  \textbf{#1}\kern1pt\up{#2}\kern1pt%
  {\tiny [#3, #4]}}

\begin{table*}[t]
\centering
\scriptsize
\setlength{\tabcolsep}{3pt}
\renewcommand{\arraystretch}{0.9}
\resizebox{\textwidth}{!}{%
\begin{tabular}{ll l l l l l l}
\toprule
\rowcolor{gray!12}
 & &
\multicolumn{2}{c}{\textbf{Response quality}} &
\multicolumn{4}{c}{\textbf{Cultural anchor tracking (\%)}} \\

\rowcolor{gray!12}
\multirow{-2}{*}{\textbf{Backbone}} &
\multirow{-2}{*}{\textbf{Method}} &
\multicolumn{1}{c}{\cellcolor{hhcol}3H Avg $\uparrow$} &
\multicolumn{1}{c}{\cellcolor{tdqcol}TDQ Avg $\uparrow$} &
\multicolumn{1}{c}{AMR $\uparrow$} &
\multicolumn{1}{c}{$\widehat{S}(5)$ $\uparrow$} &
\multicolumn{1}{c}{Contra. $\downarrow$} &
\multicolumn{1}{c}{Recovery $\uparrow$} \\
\midrule

\multicolumn{8}{c}{\cellcolor{gray!12}\textit{English}} \\
\midrule

\rowcolor{fam2}
& Base
& \cellcolor{hhcol}\baseCI{2.81}{2.76}{2.85}
& \cellcolor{tdqcol}\baseCI{2.93}{2.90}{2.96}
& \baseCI{66.6}{64.7}{68.4}
& \baseCI{30.5}{27.6}{33.5}
& \baseCI{13.2}{12.0}{14.4}
& \baseCI{55.7}{52.8}{58.7} \\

\rowcolor{fam2}
\multirow{-2}{*}{Qwen3-8B-IT}
& \textbf{Ours}
& \cellcolor{hhcol}\oursCI{3.41}{+21.4\%}{3.35}{3.46}
& \cellcolor{tdqcol}\oursCI{3.45}{+17.7\%}{3.42}{3.48}
& \oursCI{74.0}{+7.4}{72.2}{75.7}
& \oursCI{42.3}{+11.8}{39.5}{45.2}
& \oursCI{6.6}{-6.6}{5.7}{7.5}
& \oursCI{56.1}{+0.4}{52.9}{59.3} \\
\cmidrule(lr){1-8}

\rowcolor{fam3}
& Base
& \cellcolor{hhcol}\baseCI{2.53}{2.48}{2.57}
& \cellcolor{tdqcol}\baseCI{3.00}{2.97}{3.03}
& \baseCI{58.4}{56.3}{60.5}
& \baseCI{21.9}{19.5}{24.2}
& \baseCI{12.2}{10.9}{13.4}
& \baseCI{46.7}{43.8}{49.5} \\

\rowcolor{fam3}
\multirow{-2}{*}{Llama-3.1-8B-IT}
& \textbf{Ours}
& \cellcolor{hhcol}\oursCI{3.23}{+27.7\%}{3.18}{3.28}
& \cellcolor{tdqcol}\oursCI{3.31}{+10.3\%}{3.28}{3.35}
& \oursCI{75.3}{+16.9}{73.8}{76.8}
& \oursCI{53.1}{+31.2}{50.5}{56.4}
& \oursCI{7.2}{-5.0}{6.4}{8.1}
& \oursCI{62.7}{+16.0}{59.7}{65.6} \\
\cmidrule(lr){1-8}

\rowcolor{fam4}
& Base
& \cellcolor{hhcol}\baseCI{2.67}{2.63}{2.72}
& \cellcolor{tdqcol}\baseCI{3.01}{2.98}{3.04}
& \baseCI{64.0}{61.9}{66.0}
& \baseCI{25.6}{23.1}{28.2}
& \baseCI{10.9}{9.8}{12.0}
& \baseCI{49.9}{46.9}{52.8} \\

\rowcolor{fam4}
\multirow{-2}{*}{SEA-LION-v4-8B-IT}
& \textbf{Ours}
& \cellcolor{hhcol}\oursCI{3.13}{+17.2\%}{3.07}{3.18}
& \cellcolor{tdqcol}\oursCI{3.28}{+9.0\%}{3.25}{3.31}
& \oursCI{71.4}{+7.4}{69.8}{73.0}
& \oursCI{35.9}{+10.3}{33.2}{38.6}
& \oursCI{10.2}{-0.7}{9.2}{11.1}
& \oursCI{59.5}{+9.6}{56.6}{62.3} \\

\midrule
\multicolumn{8}{c}{\cellcolor{gray!12}\textit{Native}} \\
\midrule

\rowcolor{fam2}
& Base
& \cellcolor{hhcol}\baseCI{2.80}{2.75}{2.85}
& \cellcolor{tdqcol}\baseCI{2.96}{2.92}{2.99}
& \baseCI{66.5}{64.6}{68.4}
& \baseCI{26.0}{23.3}{28.8}
& \baseCI{11.2}{10.2}{12.3}
& \baseCI{54.4}{51.4}{57.6} \\

\rowcolor{fam2}
\multirow{-2}{*}{Qwen3-8B-IT}
& \textbf{Ours}
& \cellcolor{hhcol}\oursCI{3.34}{+19.3\%}{3.28}{3.39}
& \cellcolor{tdqcol}\oursCI{3.34}{+12.8\%}{3.31}{3.37}
& \oursCI{79.8}{+13.3}{78.2}{81.4}
& \oursCI{52.5}{+26.5}{49.5}{55.5}
& \oursCI{5.6}{-5.6}{4.9}{6.3}
& \oursCI{66.1}{+11.7}{62.9}{69.2} \\
\cmidrule(lr){1-8}

\rowcolor{fam3}
& Base
& \cellcolor{hhcol}\baseCI{2.54}{2.50}{2.58}
& \cellcolor{tdqcol}\baseCI{3.00}{2.97}{3.03}
& \baseCI{50.5}{48.1}{52.9}
& \baseCI{13.6}{11.8}{15.4}
& \baseCI{10.3}{9.1}{11.5}
& \baseCI{40.5}{37.6}{43.4} \\

\rowcolor{fam3}
\multirow{-2}{*}{Llama-3.1-8B-IT}
& \textbf{Ours}
& \cellcolor{hhcol}\oursCI{3.03}{+19.3\%}{2.98}{3.09}
& \cellcolor{tdqcol}\oursCI{3.17}{+5.7\%}{3.14}{3.20}
& \oursCI{76.9}{+26.4}{75.2}{78.8}
& \oursCI{40.6}{+27.0}{37.6}{43.6}
& \oursCI{7.5}{-2.8}{6.7}{8.4}
& \oursCI{63.1}{+22.6}{60.2}{66.3} \\
\cmidrule(lr){1-8}

\rowcolor{fam4}
& Base
& \cellcolor{hhcol}\baseCI{2.72}{2.68}{2.77}
& \cellcolor{tdqcol}\baseCI{3.01}{2.99}{3.04}
& \baseCI{64.3}{62.2}{66.4}
& \baseCI{25.6}{22.9}{28.3}
& \baseCI{8.8}{7.8}{9.8}
& \baseCI{50.3}{47.3}{53.3} \\

\rowcolor{fam4}
\multirow{-2}{*}{SEA-LION-v4-8B-IT}
& \textbf{Ours}
& \cellcolor{hhcol}\oursCI{3.14}{+15.4\%}{3.09}{3.20}
& \cellcolor{tdqcol}\oursCI{3.23}{+7.3\%}{3.20}{3.26}
& \oursCI{72.8}{+8.5}{71.2}{74.4}
& \oursCI{39.0}{+13.4}{36.1}{41.7}
& \oursCI{7.1}{-1.7}{6.3}{7.9}
& \oursCI{57.5}{+7.1}{54.7}{60.3} \\

\bottomrule
\end{tabular}}
\caption{Results on the \textit{CultureConverse-Long} test set.
Contra. is the fraction of applicable anchor--turn pairs
with anchor-contradicting responses.
Brackets report 95\% confidence intervals (details in App.~\ref{sec:Bootstrap confidence intervals}).
Changes from Base are relative (\%) for 3H/TDQ
and percentage points for anchor metrics.
Green/red denote improvement/degradation.
Better values within each backbone are bold.}
\label{tab:long_results}
\end{table*}

\section{Experimentation}

\subsection{Experimental Setup}

We evaluate our framework on three LLM backbones from different model families: Qwen3-8B-IT, Llama-3.1-8B-IT, and SEA-LION-v4-8B-IT. Each backbone is compared in Base, SFT, and CulSTAR configurations on the CultureConverse~\citep{tan2026cultureconverse} test set, with SFT trained on CultureConverse to represent supervised adaptation. This comparison isolates the benefit of our framework beyond standard supervised adaptation.
All variants are evaluated under the same protocol in both English and native-language interactions. Following CultureConverse, we report 1--5 Likert ratings for Helpful, Honest, Harmless, Naturalness, Scenario Plausibility, and Cultural Typicality, alongside two aggregate metrics. 3H averages Helpful, Honest, and Harmless, while Training Data Quality (TDQ) averages Naturalness, Scenario Plausibility, and Cultural Typicality to capture dialogue realism and ecological validity. Training and optimization details are provided in App.~\ref{sec:training_con}.
\begin{figure*}[htbp] \centering \begin{subfigure}[t]{0.47\linewidth} \centering \includegraphics[width=\linewidth]{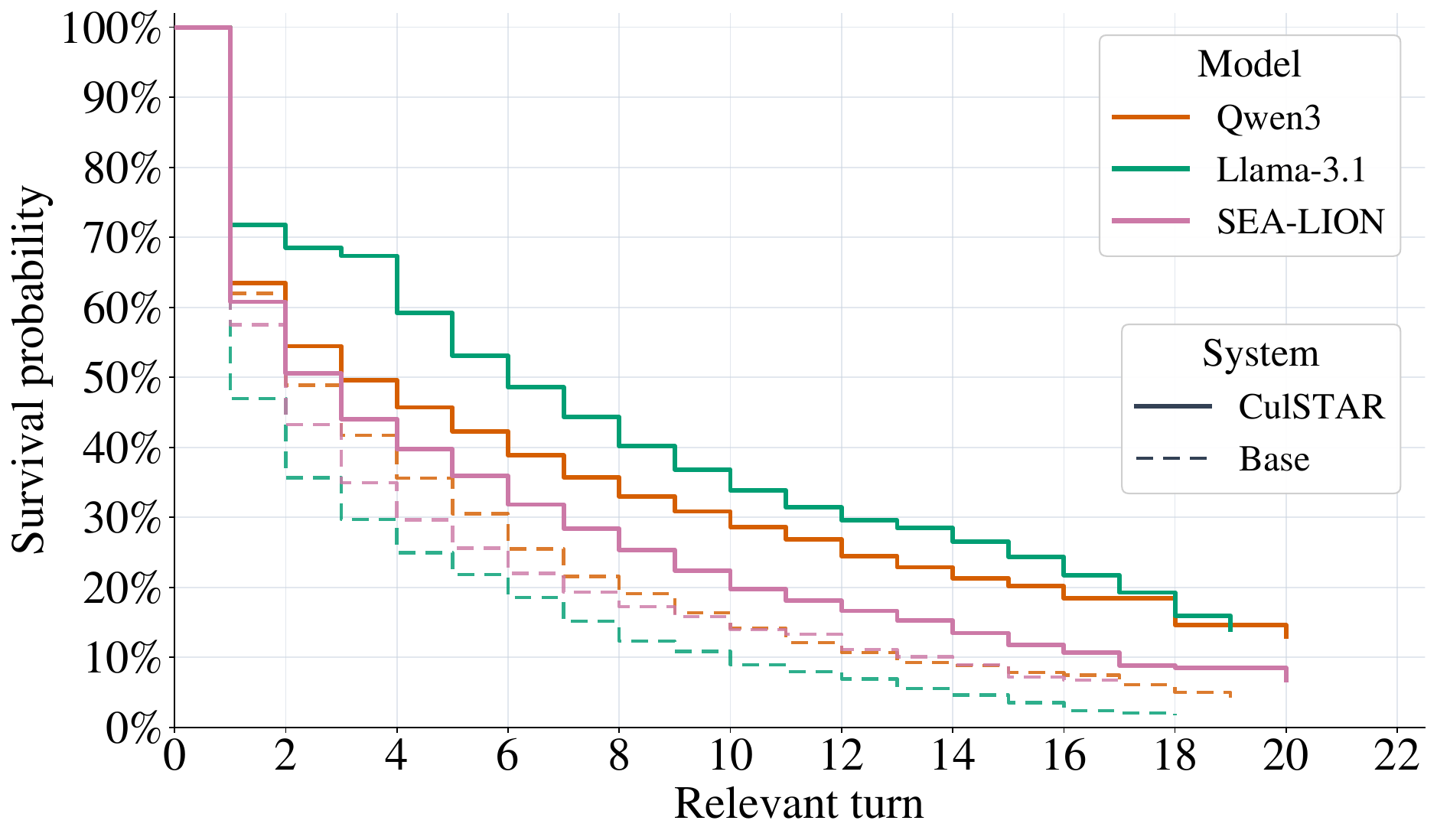} \label{fig:first} \end{subfigure} \hfill \begin{subfigure}[t]{0.47\linewidth} \centering \includegraphics[width=\linewidth]{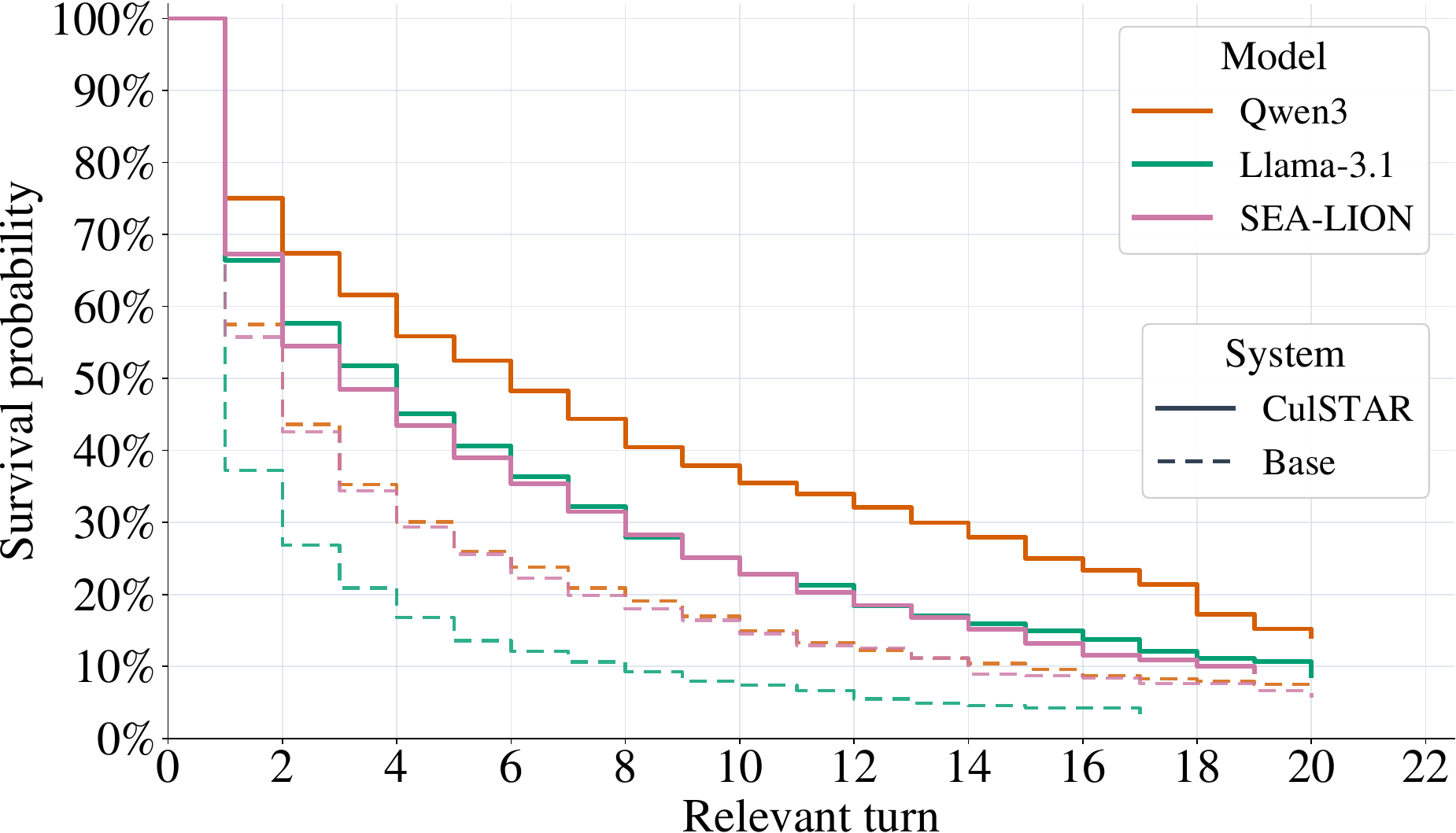} \label{fig:second} \end{subfigure} \captionsetup{skip=-8pt} \caption{Kaplan--Meier estimates of cultural-consistency survival in English (left) and native-language (right).} \label{fig:combined} \end{figure*}
\definecolor{abbase}{RGB}{246,246,246}
\definecolor{abmem}{RGB}{235,245,250}
\definecolor{abrl}{RGB}{239,236,250}
\definecolor{hhcol}{RGB}{245,241,225}
\definecolor{tdqcol}{RGB}{238,245,235}

\newcommand{\gain}[1]{\textcolor{teal!70!black}{\scriptsize #1}}
\newcommand{\dropc}[1]{\textcolor{red!70!black}{\scriptsize #1}}

\newcommand{\ablCI}[3]{%
  #1\kern1pt{\tiny [#2, #3]}}

\begin{table*}[t]
\centering
\scriptsize
\setlength{\tabcolsep}{3pt}
\renewcommand{\arraystretch}{0.9}
\resizebox{\textwidth}{!}{%
\begin{tabular}{ll l l l l l l}
\toprule

\rowcolor{gray!12}
& &
\multicolumn{2}{c}{\textbf{Response quality}}
& \multicolumn{4}{c}{\textbf{Cultural anchor tracking (\%)}} \\

\rowcolor{gray!12}
\multirow{-2}{*}{\textbf{Backbone}}
& \multirow{-2}{*}{\textbf{Configuration}}
& \multicolumn{1}{c}{\cellcolor{hhcol}3H Avg $\uparrow$}
& \multicolumn{1}{c}{\cellcolor{tdqcol}TDQ Avg $\uparrow$}
& \multicolumn{1}{c}{AMR $\uparrow$}
& \multicolumn{1}{c}{$\widehat{S}(5)$ $\uparrow$}
& \multicolumn{1}{c}{Contra. $\downarrow$}
& \multicolumn{1}{c}{Recovery $\uparrow$} \\

\midrule
\multicolumn{8}{c}{\cellcolor{gray!12}\textit{English}} \\
\midrule

\rowcolor{abbase}
&
Base
& \cellcolor{hhcol}\ablCI{2.53}{2.48}{2.57}
& \cellcolor{tdqcol}\ablCI{3.00}{2.97}{3.03}
& \ablCI{58.4}{56.3}{60.5}
& \ablCI{21.9}{19.5}{24.2}
& \ablCI{12.2}{10.9}{13.4}
& \ablCI{46.7}{43.8}{49.5} \\

\rowcolor{abmem}
&
+ DCM
& \cellcolor{hhcol}\ablCI{3.12\kern1pt \gain{(+23.3\%)}}{3.06}{3.16}
& \cellcolor{tdqcol}\ablCI{\textbf{3.31}\kern1pt \gain{(+10.3\%)}}{3.21}{3.36}
& \ablCI{73.1\kern1pt \gain{(+14.7)}}{71.4}{74.3}
& \ablCI{50.1\kern1pt \gain{(+28.2)}}{47.0}{52.4}
& \ablCI{11.2\kern1pt \gain{(-1.0)}}{9.5}{13.0}
& \ablCI{57.9\kern1pt \gain{(+11.2)}}{55.5}{60.1} \\

\rowcolor{abrl}
\multirow{-3}{*}{Llama-3.1-8B-IT}
&
+ SCCA-GRPO
& \cellcolor{hhcol}\ablCI{\textbf{3.23}\kern1pt \gain{(+27.7\%)}}{3.18}{3.28}
& \cellcolor{tdqcol}\ablCI{\textbf{3.31}\kern1pt \gain{(+10.3\%)}}{3.28}{3.35}
& \ablCI{\textbf{75.3}\kern1pt \gain{(+16.9)}}{73.8}{76.8}
& \ablCI{\textbf{53.1}\kern1pt \gain{(+31.2)}}{50.5}{56.4}
& \ablCI{\textbf{7.2}\kern1pt \gain{(-5.0)}}{6.4}{8.1}
& \ablCI{\textbf{62.7}\kern1pt \gain{(+16.0)}}{59.7}{65.6} \\

\midrule
\multicolumn{8}{c}{\cellcolor{gray!12}\textit{Native}} \\
\midrule

\rowcolor{abbase}
&
Base
& \cellcolor{hhcol}\ablCI{2.54}{2.50}{2.58}
& \cellcolor{tdqcol}\ablCI{3.00}{2.97}{3.03}
& \ablCI{50.5}{48.1}{52.9}
& \ablCI{13.6}{11.8}{15.4}
& \ablCI{10.3}{9.1}{11.5}
& \ablCI{40.5}{37.6}{43.4} \\

\rowcolor{abmem}
&
+ DCM
& \cellcolor{hhcol}\ablCI{2.93\kern1pt \gain{(+15.4\%)}}{2.87}{2.96}
& \cellcolor{tdqcol}\ablCI{3.16\kern1pt \gain{(+5.3\%)}}{3.11}{3.18}
& \ablCI{75.3\kern1pt \gain{(+24.8)}}{73.4}{76.2}
& \ablCI{38.4\kern1pt \gain{(+24.8)}}{36.6}{41.2}
& \ablCI{12.0\kern1pt \dropc{(+1.7)}}{11.1}{13.4}
& \ablCI{60.5\kern1pt \gain{(+20.0)}}{58.5}{62.4} \\

\rowcolor{abrl}
\multirow{-3}{*}{Llama-3.1-8B-IT}
&
+ SCCA-GRPO
& \cellcolor{hhcol}\ablCI{\textbf{3.03}\kern1pt \gain{(+19.3\%)}}{2.98}{3.09}
& \cellcolor{tdqcol}\ablCI{\textbf{3.17}\kern1pt \gain{(+5.7\%)}}{3.14}{3.20}
& \ablCI{\textbf{76.9}\kern1pt \gain{(+26.4)}}{75.2}{78.8}
& \ablCI{\textbf{40.6}\kern1pt \gain{(+27.0)}}{37.6}{43.6}
& \ablCI{\textbf{7.5}\kern1pt \gain{(-2.8)}}{6.7}{8.4}
& \ablCI{\textbf{63.1}\kern1pt \gain{(+22.6)}}{60.2}{66.3} \\

\bottomrule
\end{tabular}}
\caption{
Ablation of DCM and SCCA-GRPO on long-horizon
culturally grounded conversations.
+DCM: SFT warm-up only, no SCCA-GRPO.
Parentheses show changes from the corresponding Base:
relative (\%) for 3H/TDQ and percentage points
for cultural anchor metrics. Brackets report 95\% confidence intervals (details in App.~\ref{sec:Bootstrap confidence intervals}).
Green/red denote improvement/degradation.
Best values are bold.
}
\label{tab:cura_ablation}
\end{table*}

\subsection{Culturally Grounded Interaction Quality}
Table~\ref{tab:main_results_shorter} shows that CulSTAR outperforms Base and SFT in aggregate 3H and TDQ across all six backbone and language combinations, demonstrating gains beyond conventional supervised adaptation. The improvement is pronounced in native-language interactions, where the base models exhibit lower 3H scores across all three backbones. CulSTAR improves native-language 3H by 31.7\% on Llama-3.1-8B-IT and 19.2\% on SEA-LION-v4-8B-IT, together with TDQ gains of 12.1\% and 9.5\%, respectively. Clean-episode rates also improve over the corresponding base model across all settings. These consistent gains show that combining explicit cultural state maintenance with cultural credit assignment improves both general assistance quality and culturally grounded interaction beyond task-specific instruction tuning.

\subsection{Long-horizon cultural consistency}
Table~\ref{tab:long_results} shows that the advantage of CulSTAR becomes especially clear when cultural constraints must remain effective across extended interactions. Across every backbone in both English and native-language settings, CulSTAR improves anchor maintenance, five-turn survival, post-violation recovery, and response-level 3H and TDQ, while reducing cultural contradictions. The strongest gains appear on persistence metrics directly characterizing cultural drift. On Llama-3.1-8B-IT in the native-language setting, AMR increases by 26.4 percentage points and five-turn survival by 27 points, while Qwen3-8B-IT improves five-turn survival by 26.5 points and reduces contradiction from 11.2\% to 5.6\%. Figure 2 reveals the temporal structure of these gains: CulSTAR's Kaplan--Meier curves consistently show stronger cultural-consistency survival than the base models across the conversational horizon in both language settings, reflecting delayed first violations and a larger fraction of culturally consistent anchors as conversations progress. These results demonstrate that CulSTAR improves not only the quality of individual responses, but the durability of culturally grounded behavior over long conversations. Notably, CulSTAR is trained only on short conversations with reference memory, whereas evaluation uses fully self-generated memory over up to 20 turns, indicating that correct local state transitions compose into longer-horizon cultural consistency.

\section{Ablation Studies and Baseline Comparisons}
Using Llama-3.1-8B-IT as the main backbone, we conduct additional ablation studies and comparisons with other methods, primarily on the \textit{CultureConverse-Long} test set.

\begin{figure*}[t]
  \centering
  \includegraphics[width=0.98\textwidth]{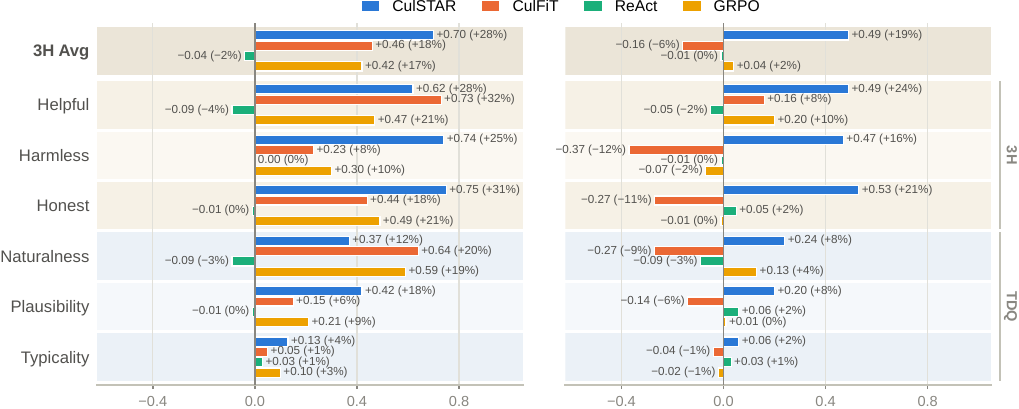}
  \captionsetup{skip=3pt} 
  \caption{Response-quality comparison of CulSTAR and representative baselines on the EN (left) and Native (right) splits. Bars show absolute score changes from the base model, with relative changes in parentheses.}
  \label{fig:likert_gain}
\end{figure*}

\begin{figure}[htbp]
  \centering
  \includegraphics[width=0.48\textwidth]{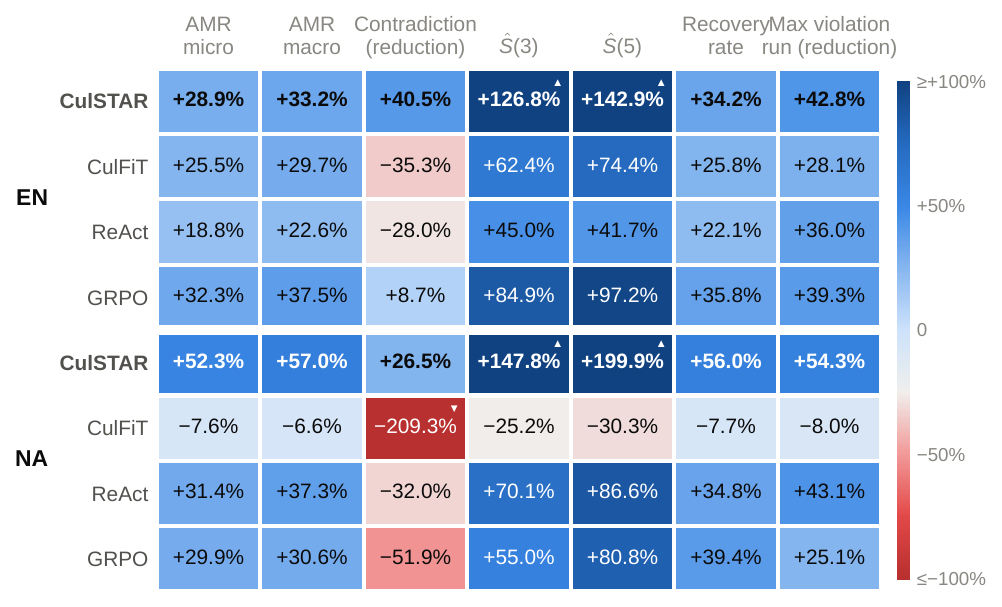}
  \captionsetup{skip=3pt} 
  \caption{Relative improvement over Llama-3.1-8B-IT on multi-turn cultural-drift
    metrics. Each cell shows
    $(\text{method} - \text{base}) / \text{base}$; positive values (blue) are
    better. For lower-is-better metrics (Contradiction, Max violation run), the
    sign is flipped: $(\text{base} - \text{method}) / \text{base}$.
    The color scale is clipped at $\pm$100\%; cells marked
    $\blacktriangle$/$\blacktriangledown$ exceed it, and the printed value is
    the actual value.
    }
  \label{fig:drift_heatmap}
\end{figure}

\subsection{Ablation Study}

Results in Tab.~\ref{tab:cura_ablation} show that Dynamic Cultural Memory and SCCA-GRPO contribute to different aspects of culturally grounded interaction. Adding Dynamic Cultural Memory produces the clearest improvements in long-horizon cultural retention, strengthening anchor maintenance and survival in both English and native-language conversations. Five-turn survival improves by 28.2 percentage points in English and AMR by 24.8 points in native-language conversations over the base model, indicating that explicitly maintaining evolving cultural state is important for preserving cultural constraints over extended interactions. Adding SCCA-GRPO on top of the memory-enhanced model further improves response-level quality and consistently reduces cross-turn contradictions. It also brings gains in anchor maintenance, survival, and recovery in both language settings. These findings suggest that stage-aware optimization primarily improves the consistent use of stored cultural information, complementing Dynamic Cultural Memory's role in sustaining cultural state. A paired bootstrap analysis further supports improvements in 3H and contradiction reduction across both language settings, with additional gains in AMR and recovery in English, as detailed in App.~\ref{sec:Bootstrap confidence intervals}.

\subsection{Comparison with Representative Baselines} We compare CulSTAR with three representative baselines that isolate distinct aspects of culturally grounded interaction. ReAct~\citep{yao2022react} represents training-free agentic prompting, interleaving reasoning and tool use to test whether agentic scaffolding alone is sufficient. CulFiT~\citep{feng2025culfit} represents culture-specific alignment, optimizing individual responses with fine-grained cultural rewards without modeling cross-turn cultural context. GRPO~\citep{shao2024deepseekmath} represents outcome-level agentic reinforcement learning, rewarding final responses without explicitly rewarding intermediate decisions that maintain cultural state. ReAct and GRPO share CulSTAR’s tools, memory design, and three-stage pipeline of response, reflection, and revision, isolating the effects of training and credit assignment. Figures~\ref{fig:likert_gain} and~\ref{fig:drift_heatmap} report changes relative to the base model. CulSTAR is the only method that improves all response-quality dimensions in both language settings. ReAct remains within 0.1 points of the base model across all dimensions, suggesting limited benefits from agentic scaffolding without learned tool-use decisions. CulFiT and GRPO are competitive in English, both outperforming CulSTAR in Naturalness. Their gains weaken sharply in native-language interactions: GRPO performs nearly on par with the base model in 3H, while CulFiT falls below it, particularly in Harmlessness and Honesty. In contrast, CulSTAR retains most of its English-language gains, indicating greater cross-language robustness.

The long-horizon metrics further distinguish anchor maintenance from contradiction avoidance. In English, all methods improve anchor maintenance, yet every baseline produces more cultural contradictions than CulSTAR. CulSTAR reduces the contradiction rate from 12.2\% to 7.2\%, compared with 11.1\% for GRPO, while ReAct and CulFiT both exceed the base-model rate. Despite slightly higher anchor maintenance, GRPO produces more contradictions than CulSTAR, showing that higher average maintenance does not necessarily prevent violations of earlier constraints. This pattern persists in native-language interactions, where CulSTAR is the only method that reduces the contradiction rate, while all baselines increase it. CulFiT underperforms the base model on all seven drift metrics, suggesting that response-level cultural alignment alone may be insufficient for sustained cultural consistency. Since CulSTAR and GRPO differ mainly in credit assignment, their performance gap suggests that rewarding intermediate decisions mainly helps avoid contradicting established constraints and improves cross-language robustness, whereas outcome-level rewards yield comparable anchor retention in English.

\section{Conclusion}
\label{sec:conclusion}
We introduced CulSTAR, combining Dynamic Cultural Memory and Stage-Aware Cultural Credit Assignment to mitigate cultural drift. We also developed CultureConverse-Long and an anchor-based longitudinal evaluation framework. Across three backbones and English and native-language interactions, CulSTAR improves response quality, anchor maintenance, and cultural-consistency survival while reducing cross-turn contradictions. These findings highlight the importance of maintaining and consistently applying evolving cultural context throughout extended interactions.

\section*{Limitations}

Although the full CulSTAR framework is evaluated across three backbones, detailed ablations and comparisons with representative baselines focus on Llama-3.1-8B-IT due to computational and API constraints. The contribution of individual training stages across additional model families remains to be examined. Our evaluation datasets primarily cover East and Southeast Asian cultural settings through simulated conversations, leaving generalization to broader cultural contexts and naturally occurring interactions open. CulSTAR also introduces additional latency, with an average runtime approximately 1.83 times that of the base model across the two datasets. Future work will extend evaluations across model families and interaction settings, refine cultural evaluation, and improve inference efficiency.

\section*{Ethical Statement}
\label{sec:ethical-statement}

CultureConverse-Long uses simulated scenarios and structured personas to evaluate culturally grounded interactions across diverse linguistic and regional contexts. Such representations inevitably simplify cultural identities and practices and may reinforce stereotypes if interpreted as universal norms or descriptions of real individuals. Cultural anchors represent scenario-specific requirements and should not be generalized into prescriptive rules for entire communities. Similarly, performance differences across regions and languages reflect model behavior under the benchmark's evaluation conditions and should not be interpreted as rankings of cultural groups.

CulSTAR maintains evolving cultural information through Dynamic Cultural Memory, including inferred attributes and contextual constraints. Incorrect cultural inferences or outdated memory entries may influence subsequent responses and propagate culturally inappropriate assumptions. Although the framework incorporates confidence estimates, supporting evidence, and memory updates, responsible deployment requires careful handling of sensitive cultural information and continued human oversight. The benchmark also includes culturally sensitive scenarios involving religious practices, social norms, and taboos, which require contextual interpretation.

Our automated evaluation relies partly on LLM judgments, which may reflect cultural biases or misclassify context-dependent behavior. We assess judgment reliability through human annotation across English and nine native languages, while acknowledging remaining disagreement and potential false positives, as detailed in Appendix~\ref{app:human evaluation}. External resources, including Wikipedia-derived knowledge, NormAd, EtiCor++, and CultureBank, are acknowledged and used in accordance with their applicable licenses. AI assistants supported manuscript editing and organization, with all content reviewed and verified by the authors. We intend to release CultureConverse-Long under the CC-BY-4.0 license for research purposes, subject to applicable third-party licensing restrictions.

\bibliography{custom}

\appendix

\section{CultureConverse-Long Dataset Construction and Statistics}
\label{app:Dataset construction}

This section describes the construction of
\textit{CultureConverse-Long}, including the original
CultureConverse methodology, our long-horizon extension,
dataset composition, cultural anchor annotations, and
comparisons with the shorter-dialogue setting.

\subsection{Dataset Construction}

\paragraph{Original CultureConverse methodology.}
CultureConverse~\citep{tan2026cultureconverse} introduces
a structured pipeline for constructing culturally grounded
multi-turn dialogues.
It first defines 58 region--identity combinations across
ten East and Southeast Asian regions, drawing on ethnicity,
religion, language, and regional background.
These identities are combined with three age cohorts
and two gender labels to form 348 persona configurations.
In parallel, ten culture-agnostic scenario templates are
generated for each of 42 subdomains spanning seven domains,
yielding 420 templates.
Pairing personas with templates produces 146,160 scenario
seeds, which are assigned to standard, contextual-friction,
and normative-friction challenge categories.

Each seed is subsequently expanded into a scenario blueprint
grounded in retrieved cultural knowledge and taboo evidence.
The original framework uses a Wikipedia-derived cultural
knowledge base containing 180,992 entries and a curated taboo
knowledge base containing 962 entries.
Each blueprint specifies public user information, private
simulator context, and oracle-level cultural constraints,
including relevant norms, taboos, and success or failure
criteria.
This separation ensures that the evaluated assistant must
infer hidden cultural requirements from the interaction.
Finally, each blueprint is decomposed into 3--5 ordered
interaction shards specifying user objectives, disclosure
instructions, and success or failure transitions.
These shards guide an interactive user simulator that
progressively reveals contextual information while interacting
with the evaluated assistant.

\paragraph{Generation models and quality assessment.}
We follow CultureConverse~\citep{tan2026cultureconverse}
in selecting models for data generation and quality assessment.
Specifically, we employ GPT-5.2 for scenario template generation
and GPT-5 mini for blueprint construction, shard generation,
user simulation, and automated quality evaluation.
We also adopt its evaluation dimensions, including naturalness,
scenario plausibility, and cultural typicality.
Notably, the original study has already validated its
model-based evaluator against human annotations, achieving
90.1\% within-one agreement and a mean absolute error of
0.581, supporting the reliability of the adopted
evaluation protocol.

\paragraph{Long-horizon extension.}
Following this methodology, we construct
\textit{CultureConverse-Long} to support the evaluation
of cultural consistency over extended interactions.
Each scenario contains 10--20 interaction shards,
with an average of 13.64, supporting conversations
of up to 20 turns.
Each scenario retains the structured blueprint and
shard-level objectives of CultureConverse, including cultural
norms, taboos, and success and failure criteria.
The extended shard sequences additionally specify cross-turn
recall requirements, allowing constraints introduced at
earlier stages to become relevant again later.
This design supports the evaluation of recurring cultural
requirements over longer interaction histories.

\paragraph{Dialogue simulation.}
Following CultureConverse, the simulator uses the scenario
blueprint and active shard to guide the user's interaction
with the evaluated assistant.
Cultural information is disclosed progressively, while the assistant receives only the user-facing conversation history, with the hidden scenario specification withheld.
Each scenario is evaluated separately in English and its
corresponding native language.
The resulting dialogue trajectories provide the basis for
turn-level cultural consistency evaluation.

\begin{figure*}[t]
  \centering
  \includegraphics[width=\textwidth]{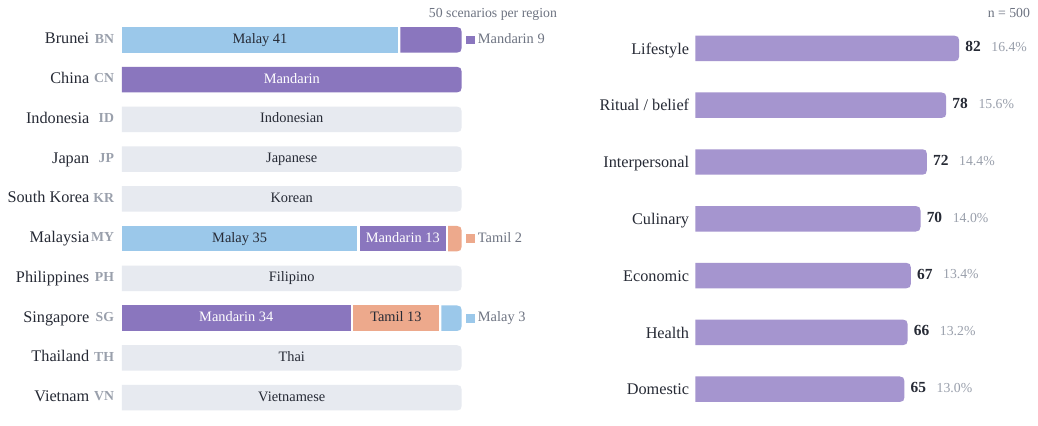}
  \caption{\textbf{CultureConverse-Long dataset composition.}
Left: Native-language distribution across ten regions,
each containing 50 scenarios.
Right: Scenario distribution across seven cultural domains.
Each scenario is evaluated in both English and its
corresponding native language.}
\label{fig:dataset_composition}
\end{figure*}

\subsection{Dataset Composition}

\paragraph{Regional and linguistic coverage.}
CultureConverse-Long contains 500 unique scenarios
distributed evenly across ten East and Southeast Asian
regions, with 50 scenarios per region.
The scenarios cover 58 user-identity combinations and
nine native languages.
Each scenario is evaluated in both English and its
corresponding native language, yielding 1,000 evaluation
episodes per model across ten languages.
Figure~\ref{fig:dataset_composition} summarizes the regional
and native-language distribution.

\paragraph{Scenario diversity.}
The dataset spans seven cultural domains and 42 subdomains,
covering everyday interactions involving social relationships,
religious practices, professional conduct, domestic life,
mobility, food, and health.
Figure~\ref{fig:dataset_composition} presents the distribution
across domains.
The sampled personas are approximately balanced by gender,
with 251 female and 249 male personas.
The three age cohorts (18--34, 35--54, and 55+) account
for 169, 181, and 150 scenarios, respectively.

\paragraph{Challenge configurations.}
Following the challenge taxonomy of CultureConverse,
we organize scenarios into three categories:
Standard, Contextual Friction, and Normative Friction.
Standard scenarios focus on routine culturally grounded
assistance, whereas contextual and normative friction
introduce cross-cultural tensions or culturally sensitive
constraints.
The dataset contains 259 standard scenarios (51.8\%),
130 contextual-friction scenarios (26.0\%), and
111 normative-friction scenarios (22.2\%).
Additionally, 130 scenarios (26.0\%) involve cross-cultural
interactions, and 111 (22.2\%) are grounded in cultural taboos.
These configurations broaden the range of cultural
requirements encountered during extended interactions.

\subsection{Cultural Anchor Construction and Statistics}

\paragraph{Anchor specification.}
To assess cultural consistency beyond individual responses,
we derive cultural anchors from the scenario blueprints
and shard-level specifications.
Each anchor represents a cultural requirement associated
with the scenario, including relevant norms, taboos,
situational constraints, or interaction expectations.
Anchors are aligned with the dialogue turns at which
their requirements are applicable, enabling evaluation
of whether previously established cultural constraints
remain satisfied as interactions progress.

We further organize anchors along two dimensions:
core versus peripheral, and global versus local.
These annotations capture differences in the importance
and contextual scope of cultural requirements.
The resulting anchor annotations support the analysis of
cultural maintenance, attenuation, replacement,
contradiction, and the first observed violation
of an established constraint.

\paragraph{Anchor composition.}
CultureConverse-Long contains 3,090 cultural anchors,
averaging 6.18 per scenario, with a median of 5 and
a maximum of 17.
Core and peripheral anchors account for 67.1\% and
32.9\%, respectively, while global and local anchors
account for 64.7\% and 35.3\%.
These annotations provide a structured basis for
examining cultural requirements at different levels
of importance and contextual scope.

\subsection{Comparison with Shorter Dialogues}

To characterize the long-horizon extension, we compare
CultureConverse-Long with a 500-scenario shorter-dialogue
reference set using the same cultural anchor statistics.
This comparison focuses on the number, composition,
and temporal recurrence of cultural constraints.
Table~\ref{tab:long_short_compare} summarizes
the principal differences.

\begin{table*}[t]
\centering
\small
\setlength{\tabcolsep}{9pt}
\renewcommand{\arraystretch}{1.08}
\begin{tabular}{lcc}
\toprule
\textbf{Metric} &
\textbf{Short} &
\textbf{CultureConverse-Long} \\
\midrule
\multicolumn{3}{l}{\textit{Dataset and anchor composition}} \\
\midrule
Scenarios                            & 500     & 500 \\
Cultural anchors                    & 2,607   & 3,090 \\
Average anchors per scenario        & 5.21    & 6.18 \\
Median / maximum anchors            & 5 / 9   & 5 / 17 \\
Core / peripheral anchors           & 63.8 / 36.2\% & 67.1 / 32.9\% \\
Global / local anchors              & 57.3 / 42.7\% & 64.7 / 35.3\% \\
\midrule
\multicolumn{3}{l}{\textit{Designed cross-shard recurrence}} \\
\midrule
Recurrent anchors ($\geq 2$ shards)
                                    & 1,314   & 1,854 \\
Recurrent anchor ratio              & 50.4\%  & \textbf{60.0\%} \\
Mean first-revisit distance (shards)
                                    & --      & 2.36 \\
Median / P90 first-revisit distance
                                    & --      & 2 / 5 \\
Mean first-to-last span (shards)    & --      & 7.72 \\
Median / maximum first-to-last span & --      & 8 / 18 \\
First-revisit distance $\geq 5$ shards
                                    & --      & 11.1\% \\
\bottomrule
\end{tabular}
\caption{Comparison of cultural anchor composition
and designed recurrence between the shorter-dialogue
reference set and CultureConverse-Long.
Recurrent anchors are those designed to appear in at
least two distinct interaction shards.
Revisit-distance and span statistics are computed
over recurrent anchors. A dash (--) indicates that the corresponding statistic is not applicable.}
\label{tab:long_short_compare}
\end{table*}

\paragraph{Anchor density and recurrence.}
The long-horizon extension increases the total number
of cultural anchors from 2,607 to 3,090, corresponding
to an increase in average anchors per scenario from
5.21 to 6.18.
More importantly, the proportion of anchors designed
to recur across at least two interaction shards increases
from 50.4\% to 60.0\%.
This higher recurrence coverage provides more opportunities
to assess whether cultural requirements introduced at
earlier interaction stages are appropriately maintained
when they become relevant again.

\paragraph{Designed recurrence distances.}
We further characterize the temporal structure of the
1,854 recurrent anchors in CultureConverse-Long.
The mean distance between the first designed occurrence
and its first subsequent revisit is 2.36 shards
(median 2; 90th percentile 5).
The mean span from the first to the final designed
occurrence reaches 7.72 shards (median 8), with
a maximum of 18.
Furthermore, 24.3\% of recurrent anchors have a
first-revisit distance of at least three shards,
and 11.1\% have a distance of at least five shards.

The first-revisit distance captures the initial
separation between repeated occurrences, whereas
the first-to-last span reflects how far a cultural
requirement extends across the planned interaction.
Importantly, this span measures the separation between
designed occurrences and does not imply that the
constraint is continuously applicable at every
intermediate shard.

\subsection{Realized Long-Horizon Dialogue Characteristics}

\paragraph{Turn-level recurrence.}
Designed shard sequences do not necessarily correspond
one-to-one with realized dialogue turns.
We therefore additionally examine anchor recurrence
in generated dialogue trajectories.
Across 12 model, system, and language conditions,
the mean first-revisit distance ranges from 2.54
to 3.16 turns, with a median of two turns across
all conditions.
The 90th percentile ranges from five to six turns.
First-revisit distances of at least five turns
account for 13.0--20.3\% of recurrent anchors,
while distances of at least ten turns account
for 0.7--3.6\%.
These statistics characterize the realized temporal
separation of recurring cultural requirements
under different interaction conditions.

\paragraph{Multi-turn anchor applicability.}
We also examine the number of dialogue turns at which
each anchor is applicable.
In CultureConverse-Long, 67.9--74.3\% of anchors
are applicable for at least two turns, and
56.5--65.5\% are applicable for at least five turns
across evaluation conditions.
By comparison, the corresponding ranges in the
shorter-dialogue setting are 66.3--69.5\% and
25.4--39.2\%, respectively.
Table~\ref{tab:long_applicability} summarizes
these results.

\begin{table}[t]
\centering
\small
\setlength{\tabcolsep}{5pt}
\renewcommand{\arraystretch}{1.05}
\begin{tabular}{lcc}
\toprule
\textbf{Applicable turns} &
\textbf{Short} &
\textbf{Long} \\
\midrule
$\geq 2$ turns & 66.3--69.5\% & 67.9--74.3\% \\
$\geq 5$ turns & 25.4--39.2\% & 56.5--65.5\% \\
\bottomrule
\end{tabular}
\caption{Fraction of cultural anchors applicable
at multiple dialogue turns in shorter- and
longer-dialogue settings.
Ranges summarize the observed values across
evaluation conditions.
Applicability indicates opportunities for evaluation,
not successful constraint maintenance.}
\label{tab:long_applicability}
\end{table}

The increased coverage of anchors applicable at
five or more turns provides a broader basis for
evaluating cultural consistency over extended
interactions.
Together with the designed cross-shard recurrence
statistics, these results characterize
CultureConverse-Long as an evaluation setting
with recurring cultural requirements distributed
across longer dialogue histories.

\subsection{Detailed Identity Coverage}
\label{app:identity_coverage}

Table~\ref{tab:ccl-6} provides the complete distribution
of region--identity combinations in CultureConverse-Long.
Although the dataset assigns 50 scenarios to each of the
ten regions, the number of scenarios varies across identity
groups within each region.
We preserve these distinctions to characterize the
cultural diversity represented in the evaluation set.
The assigned native languages reflect the dataset
configuration and do not imply uniform language use
within an identity group.

\begin{table*}[htbp] \centering \small \setlength{\tabcolsep}{5pt} \renewcommand{\arraystretch}{1.0} \begin{tabular}{@{}ll lrr@{}} \toprule \textbf{Region} & \textbf{Identity combination} & \textbf{Assigned native language} & \textbf{N} & \textbf{\%} \\ \midrule BN & Brunei Malay Muslim & Malay & 41 & 8.2 \\ BN & Brunei Chinese Christian & Mandarin & 6 & 1.2 \\ BN & Brunei Chinese Buddhist & Mandarin & 3 & 0.6 \\ \addlinespace[2pt] CN & Han Traditional Rural & Mandarin & 26 & 5.2 \\ CN & Uyghur Muslim & Mandarin & 19 & 3.8 \\ CN & Han Secular Urban & Mandarin & 2 & 0.4 \\ CN & Han Christian Protestant & Mandarin & 1 & 0.2 \\ CN & Hui Muslim & Mandarin & 1 & 0.2 \\ CN & Tibetan Mongolian Buddhist & Mandarin & 1 & 0.2 \\ \addlinespace[2pt] ID & Javanese Muslim Abangan & Indonesian & 12 & 2.4 \\ ID & Javanese Muslim Santri & Indonesian & 12 & 2.4 \\ ID & Chinese Indonesian Buddhist/Confucian & Indonesian & 9 & 1.8 \\ ID & Balinese Hindu & Indonesian & 8 & 1.6 \\ ID & Chinese Indonesian Christian & Indonesian & 4 & 0.8 \\ ID & Sundanese Muslim & Indonesian & 2 & 0.4 \\ ID & Batak Christian & Indonesian & 1 & 0.2 \\ ID & Eastern Indonesia Christian & Indonesian & 1 & 0.2 \\ ID & Minangkabau Muslim & Indonesian & 1 & 0.2 \\ \addlinespace[2pt] JP & Ryukyuan Okinawan & Japanese & 30 & 6.0 \\ JP & Yamato Buddhist & Japanese & 8 & 1.6 \\ JP & Ainu Indigenous & Japanese & 6 & 1.2 \\ JP & Zainichi Korean & Japanese & 4 & 0.8 \\ JP & Yamato Shinto & Japanese & 2 & 0.4 \\ \addlinespace[2pt] KR & Korean Christian Protestant & Korean & 24 & 4.8 \\ KR & Korean Secular & Korean & 21 & 4.2 \\ KR & Korean Buddhist & Korean & 2 & 0.4 \\ KR & Korean Christian Catholic & Korean & 2 & 0.4 \\ KR & Jeju Islander & Korean & 1 & 0.2 \\ \addlinespace[2pt] MY & Borneo Bumiputera Christian & Malay & 17 & 3.4 \\ MY & Borneo Bumiputera Muslim & Malay & 16 & 3.2 \\ MY & Chinese Christian & Mandarin & 12 & 2.4 \\ MY & Malay Muslim & Malay & 2 & 0.4 \\ MY & Chinese Taoist/Buddhist & Mandarin & 1 & 0.2 \\ MY & Indian Muslim & Tamil & 1 & 0.2 \\ MY & Tamil Hindu & Tamil & 1 & 0.2 \\ \addlinespace[2pt] PH & Visayan Catholic & Filipino & 16 & 3.2 \\ PH & Moro Muslim & Filipino & 15 & 3.0 \\ PH & Filipino INC & Filipino & 14 & 2.8 \\ PH & Filipino Evangelical & Filipino & 2 & 0.4 \\ PH & Chinoy & Filipino & 1 & 0.2 \\ PH & Ilocano Catholic & Filipino & 1 & 0.2 \\ PH & Tagalog Catholic & Filipino & 1 & 0.2 \\ \addlinespace[2pt] SG & Chinese Christian & Mandarin & 24 & 4.8 \\ SG & Indian Muslim & Tamil & 8 & 1.6 \\ SG & Chinese Secular FreeThinker & Mandarin & 7 & 1.4 \\ SG & Tamil Hindu & Tamil & 5 & 1.0 \\ SG & Chinese Taoist/Buddhist & Mandarin & 3 & 0.6 \\ SG & Malay Muslim Sunni Shafii & Malay & 3 & 0.6 \\ \addlinespace[2pt] TH & Northern Lanna & Thai & 27 & 5.4 \\ TH & Thai Malay Muslim & Thai & 18 & 3.6 \\ TH & Thai Chinese & Thai & 3 & 0.6 \\ TH & Thai Buddhist Central & Thai & 1 & 0.2 \\ TH & Thai Isan & Thai & 1 & 0.2 \\ \addlinespace[2pt] VN & Kinh Catholic & Vietnamese & 12 & 2.4 \\ VN & Kinh Secular Ancestor Worship & Vietnamese & 12 & 2.4 \\ VN & Kinh Buddhist Mahayana & Vietnamese & 10 & 2.0 \\ VN & Khmer Krom Buddhist & Vietnamese & 9 & 1.8 \\ VN & Hoa Chinese & Vietnamese & 7 & 1.4 \\ \midrule \multicolumn{3}{l}{\textbf{Total}} & \textbf{500} & \textbf{100.0} \\ \bottomrule \end{tabular} \caption{\textbf{Detailed region--identity distribution
of CultureConverse-Long.}
Counts refer to unique scenarios, with percentages computed
over all 500 scenarios. Each region contains 50 scenarios.
Native-language assignments follow the dataset configuration
and do not imply uniform language use within identity groups.
Region abbreviations: BN (Brunei), CN (China),
ID (Indonesia), JP (Japan), KR (South Korea),
MY (Malaysia), PH (Philippines), SG (Singapore),
TH (Thailand), and VN (Vietnam).} \label{tab:ccl-6} \end{table*}

\section{SCCA-GRPO Reward Details}
\label{app:reward_details}

Before SCCA-GRPO training, we perform supervised fine-tuning (SFT) on single-turn examples (Table~\ref{tab:training-config}) to stabilize the output format. This section details the reward components and weighting coefficients used by SCCA-GRPO. The coefficients are summarized in Table~\ref{tab:reward_coefficients}.

\begin{table*}[htbp]
\centering
\small
\setlength{\tabcolsep}{4.5pt}
\renewcommand{\arraystretch}{0.95}
\begin{tabular}{lll}
\toprule
\textbf{Objective} & \textbf{Component} & \textbf{Coefficient} \\
\midrule

\multirow{9}{*}{Response (rule)}
& Output validity & 0.0205 \\
& High-level planning & 0.0308 \\
& Tool selection & 0.1027 \\
& Retrieval-query coverage & 0.0616 \\
& Language assistance & 0.0205 \\
& Memory capture & 0.0514 \\
& Memory persistence & 0.0411 \\
& Memory-value correctness & 0.0308 \\
& Clarification decision & 0.0205 \\

\midrule

\multirow{6}{*}{Response (judge)}
& Goal completion & 0.1118 \\
& Cultural consistency & 0.1229 \\
& Taboo avoidance & 0.0671 \\
& Clarification quality & 0.0447 \\
& Language adequacy & 0.0894 \\
& Memory-update correctness & 0.1341 \\

\midrule

\multirow{3}{*}{Response penalties}
& Hallucination penalty & $0.10$ \\
& Unnecessary tool calls
& $-0.05$ each, capped at $-0.15$ \\
& Newly introduced memory duplicates
& $-0.03$ each, capped at $-0.09$ \\

\midrule

\multirow{5}{*}{Reflection}
& Schema-validity gate
& Multiplicative, $s_{\mathrm{schema}}\in[0,1]$ \\
& Revision-required correctness ($\alpha_1$) & 0.45 \\
& Issue-category F1 ($\alpha_2$) & 0.25 \\
& Issue-content quality ($\alpha_3$) & 0.30 \\
& Over-flagging penalty
& $-0.05$ each, capped at $-0.15$ \\

\midrule

\multirow{5}{*}{Revision}
& Revised-response quality ($\beta_1$) & 0.40 \\
& Reflection-issue resolution ($\beta_2$) & 0.25 \\
& Reference alignment ($\beta_3$) & 0.15 \\
& Asymmetric improvement score ($\beta_4$) & 0.20 \\
& Hallucination penalty ($\beta_5$) & $0.10$ \\

\midrule

\multirow{6}{*}{Floor penalty}
& Tool selection $(f_m,c_m)$ & $(0.05,0.20)$ \\
& Retrieval-query coverage $(f_m,c_m)$ & $(0.35,0.15)$ \\
& Memory capture $(f_m,c_m)$ & $(0.50,0.20)$ \\
& Memory persistence $(f_m,c_m)$ & $(0.03,0.25)$ \\
& Memory-value correctness $(f_m,c_m)$ & $(0.03,0.15)$ \\
& Memory-update correctness $(f_m,c_m)$ & $(0.30,0.20)$ \\

\bottomrule
\end{tabular}
\caption{
Reward coefficients used in SCCA-GRPO.
The positive response reward budgets are approximately
$0.38$ for rule-based components and $0.57$ for
judge-based components.
Response and reflection penalties are applied
as specified in their respective reward definitions.
The reflection reward is multiplicatively gated
by schema validity.
For the floor penalty, $(f_m,c_m)$ denotes the
component-specific threshold and penalty coefficient,
respectively.
The asymmetric improvement score is defined in
Eq.~\ref{eq:improvement_reward}.
All final task rewards are clipped to $[0,1]$.
}
\label{tab:reward_coefficients}
\end{table*}

\subsection{Reward Component Definitions}
\label{sec:Reward Component Definitions}

\paragraph{Response reward.}
We instantiate the rule-based and judge-based response rewards as separate
weighted sums. The rule-based reward is defined as
\begin{equation}
R_{\mathrm{rule}}
=
\sum_{i\in\mathcal{I}_{\mathrm{rule}}}
w_i^{\mathrm{rule}} r_i,
\label{eq:rule_reward}
\end{equation}
while the judge-based reward is
\begin{equation}
R_{\mathrm{judge}}
=
\sum_{j\in\mathcal{I}_{\mathrm{judge}}}
w_j^{\mathrm{judge}} r_j.
\label{eq:judge_reward}
\end{equation}

The deterministic components in Eq.~\ref{eq:rule_reward} cover output validity,
high-level planning, tool selection, retrieval-query coverage, language
assistance, memory capture, memory persistence, memory-value correctness, and
clarification. The LLM-judge components in Eq.~\ref{eq:judge_reward} evaluate
goal completion, cultural consistency, taboo avoidance, clarification quality,
language adequacy, and memory-update correctness.

For retrieval, query coverage combines cultural-entity coverage with relevance
to the current cultural topic. Memory supervision is computed from the true
pre-turn state and evaluates only the state transition induced at the current
turn. Specifically, we separately score identity-level memory capture, the
field--attribute structure of the induced memory delta, and the correctness
of stored values.

We define the overall response penalty as
\begin{equation}
P_{\mathrm{resp}}
=
P_{\mathrm{tool}}
+
P_{\mathrm{dup}}
+
P_{\mathrm{hall}}^{\mathrm{resp}}
+
P_{\mathrm{floor}},
\label{eq:response_penalty}
\end{equation}
where the unnecessary-tool penalty is defined as
\begin{equation}
P_{\mathrm{tool}}
=
\min\left(
0.05n_{\mathrm{tool}},
0.15
\right),
\label{eq:tool_penalty}
\end{equation}
and the duplicate-memory penalty is defined as
\begin{equation}
P_{\mathrm{dup}}
=
\min\left(
0.03n_{\mathrm{dup}},
0.09
\right).
\label{eq:duplicate_penalty}
\end{equation}

In Eqs.~\ref{eq:tool_penalty} and~\ref{eq:duplicate_penalty},
$n_{\mathrm{tool}}$ denotes the number of unnecessary tool calls, while
$n_{\mathrm{dup}}$ denotes the number of newly introduced duplicate memory
entries. The term $P_{\mathrm{hall}}^{\mathrm{resp}}\in[0,0.5]$ in
Eq.~\ref{eq:response_penalty} penalizes unsupported claims, while
$P_{\mathrm{floor}}$ imposes additional penalties when selected process
dimensions fall below their prescribed minimum levels.

For selected process dimensions, we additionally apply
\begin{equation} P_{\mathrm{floor}} = \sum_{m\in\mathcal{F}} c_m \max(0,f_m-r_m), \label{eq:p_floor} 
\end{equation}
where $r_m$ is the corresponding component score, $f_m$ is its minimum
desired floor, and $c_m$ controls the penalty applied below that floor.

\paragraph{Reflection reward.}
The structural gate
$s_{\mathrm{schema}}\in[0,1]$ evaluates whether the reflection follows the
required structured schema. The output contains four issue lists together
with a Boolean \texttt{revision\_required} decision. A valid Boolean contributes
half of the schema score, while the remaining half is assigned proportionally
to the validity of the four issue lists. An unparsable reflection receives
zero reward.

$R_{\mathrm{required}}$ measures the correctness of the
\texttt{revision\_required} decision. $R_{\mathrm{category}}$ is the F1 score
between predicted and reference active issue categories.
$R_{\mathrm{content}}$ evaluates whether the reported issues are genuine and
materially relevant. Overflagging is penalized as
\begin{equation} P_{\mathrm{overflag}} = \min(0.05n_{\mathrm{extra}},0.15), \label{eq:p_overflag} \end{equation}
where $n_{\mathrm{extra}}$ denotes the number of predicted issue categories
absent from the reference reflection.

\paragraph{Revision reward.}
A comparative judge jointly evaluates
the original response, the independent reflection, the reference revision,
and the proposed revision. It produces the revised-response quality
$Q_{\mathrm{new}}$, reflection-resolution score $R_{\mathrm{resolve}}$,
reference-alignment score $R_{\mathrm{align}}$, and hallucination penalty
$P_{\mathrm{hall}}$.

To quantify whether revision improves upon the original response, we define
$\Delta Q$ as the change in response quality from the original response to the
revised response:
\begin{equation}
\Delta Q
=
Q_{\mathrm{new}}
-
Q_{\mathrm{old}},
\label{eq:quality_difference}
\end{equation}
where $Q_{\mathrm{old}}$ and $Q_{\mathrm{new}}$ denote the quality scores of
the original and revised responses, respectively. Thus, a positive
$\Delta Q$ indicates an improvement after
revision, whereas a negative value indicates a quality regression.

We transform this quality difference using the asymmetric shaping function
\begin{equation}
g(\Delta Q)
=
\begin{cases}
\min(\Delta Q,0.5),
& \Delta Q\geq0,\\
\max(1.5\Delta Q,-0.5),
& \Delta Q<0,
\end{cases}
\label{eq:quality_shaping}
\end{equation}
and define the resulting comparative improvement reward as
\begin{equation}
R_{\Delta}
=
0.5+g(\Delta Q).
\label{eq:improvement_reward}
\end{equation}

This asymmetric shaping assigns a larger penalty to a quality regression than
the reward assigned to an improvement of the same magnitude. It therefore
encourages revision to correct identified issues while preserving already
valid content.

\subsection{Design Rationale and Coefficient Selection}
\label{app:coefficient_selection}

The reward coefficients in Table~\ref{tab:reward_coefficients} are initialized according to functional priorities and calibrated using rollout statistics from the SFT policy on a held-out development split. The design follows the principles below.

\paragraph{Normalization and fixed budgets.}
Every component is scaled to $[0,1]$. Within the response reward, rule-based components share
a total weight of approximately $0.38$ and judge-based components share $0.57$, allocating roughly $40\%$ of the positive reward budget to verifiable process behavior and $60\%$ to semantic outcome quality. The initial weights within each group reflect the functional importance of individual components, while the group-level budgets remain fixed during calibration.

\paragraph{Variance-aware weighting.}
We evaluate the SFT policy on the development split
using the same turn-level rollout protocol as SCCA-GRPO, in which each instance is conditioned on the reference pre-turn memory,
examining component saturation and within-group reward
variation.
Components that are already saturated after SFT,
such as output validity, show little within-group
variance and receive low weights ($0.021$).
Components that provide informative within-group
variation and are critical to agent function receive
higher weights, including tool selection ($0.103$),
retrieval-query coverage ($0.062$), and the memory
components.
Memory-update correctness receives the largest judge
weight ($0.134$), since a memory error affects
subsequent turns.
The calibration redistributes weights within each
group while preserving the fixed reward budgets.

\paragraph{Gate-first reflection.}
In the reflection reward, the
\texttt{revision\_required} decision has the largest
weight ($\alpha_1=0.45$), reflecting its role in
determining whether revision is triggered.
Category accuracy ($\alpha_2$) and content accuracy
($\alpha_3$) then reward correct diagnosis.

\paragraph{Asymmetric revision.}
Revised-response quality ($\beta_1$) has the largest
revision weight, and $R_{\Delta}$ applies a
$1.5$-fold stronger penalty slope to quality
regressions before clipping
(Eq.~\ref{eq:quality_shaping}).
This reflects the deployment cost: turning a correct
response into a worse one is more harmful than leaving
it unrevised.

\paragraph{Regression floors.}
The floor thresholds $f_m$ are calibrated from
step-0 validation rollouts of the SFT policy
on tool selection, retrieval-query coverage,
memory capture, memory persistence,
memory-value correctness, and
memory-update correctness.

The floor penalty (Eq.~\ref{eq:p_floor}) is active
only when a component falls below its corresponding
baseline threshold and gives no bonus above it.
It therefore discourages trading structural abilities
for gains in fluency-related judge scores.

\paragraph{Anti-hacking costs.}
Capped costs on unnecessary tool calls
(Eq.~\ref{eq:tool_penalty}), duplicate memory additions
(Eq.~\ref{eq:duplicate_penalty}), and over-flagging
in reflection (Eq.~\ref{eq:p_overflag}) target three
shortcuts: calling every tool, repeatedly adding
existing memory entries, and reporting issues for
every response.
The caps keep these costs from dominating the reward.

\paragraph{Reward composition.}
The resulting reward provides stage-aware,
process-informed training signals.
Deterministic credits are derived from verifiable
trajectory annotations, semantic credits are supplied
by an independent LLM judge, and the composite reward
is converted into a group-relative advantage for
policy optimization.
The complete reward configuration, including component
weights, floor thresholds, and penalty coefficients,
is finalized on the development split and frozen
before test-set evaluation.
The same configuration is applied to Qwen, Llama,
and SEA-LION without backbone-specific retuning.

\section{Detailed Cultural Drift Evaluation}
\label{app:cultural_drift}

\subsection{Cultural Anchors}

We evaluate cultural drift as a temporal failure to preserve previously established cultural constraints during a multi-turn conversation. The evaluation unit is an anchor--turn pair. A cultural anchor represents a self-contained and verifiable cultural requirement, such as a taboo, ritual norm, culturally appropriate role relationship, or context-specific behavioral constraint.

Anchors are derived from the scenario blueprint, including its cultural constraints, safety tripwires, and shard-level success and failure conditions. Each anchor is marked as either global, meaning that it remains valid across multiple dialogue stages, or local, meaning that it applies only to a particular shard or task stage. Anchors are additionally categorized as core or peripheral, depending on whether violating them undermines the central cultural premise of the scenario. To ensure comparability, the same cached anchors and anchor-to-turn mappings are shared across all evaluated systems.

\subsection{Two-Stage Anchor--Turn Evaluation}

For every assistant turn $t$ and candidate cultural anchor $a$, we first determine whether the anchor is relevant to that turn. If the dataset explicitly maps the anchor to the current turn, the pair is directly marked as applicable. Otherwise, an LLM judge performs a response-blind applicability assessment using the anchor proposition, its source evidence, the current user message, and the minimum dialogue history required to resolve references or previously established commitments. The assistant response is deliberately excluded during this stage, preventing a system from avoiding evaluation simply by ignoring a cultural requirement that should have been considered.

We first determine whether anchor $a$ is applicable at turn $t$. Only applicable anchor--turn pairs are included in the cultural-consistency evaluation.
In the second stage, the judge observes the assistant response and assigns it one of five mutually exclusive states $Y_{a,t}$.
\texttt{EXPLICIT} indicates that the response explicitly and correctly applies the anchor, whereas \texttt{IMPLICIT} indicates that the response remains compatible with the anchor without restating it. These two states form the \textbf{MAINTAINED} set $\mathcal{S}_{\mathrm{M}}$.
\texttt{ATTENUATED} indicates that the anchor is weakened, omitted, or generalized despite being relevant; \texttt{REPLACED} indicates that the original cultural framework is substituted by a different one; and \texttt{CONTRADICTED} indicates that the response directly reverses or violates the anchor. These three states form the \textbf{VIOLATED} set $\mathcal{S}_{\mathrm{V}}$.
The binary maintenance indicator is
\begin{equation}
    M_{a,t} = \mathds{1}\!\left[\, Y_{a,t} \in \mathcal{S}_{\mathrm{M}} \,\right],
\end{equation}
so $M_{a,t}=0$ whenever $Y_{a,t} \in \mathcal{S}_{\mathrm{V}}$.

\subsection{Anchor Maintenance Rate}

Let $\mathcal{A}$ denote the set of canonical anchors with at least one applicable turn, and let $\mathcal{P}$ denote the set of applicable anchor--turn pairs. The primary turn-level metric is the micro-averaged Anchor Maintenance Rate:
\begin{equation}
    \mathrm{AMR}_{\mathrm{micro}}
    =
    \frac{
        \sum_{(a,t)\in\mathcal{P}} M_{a,t}
    }{
        |\mathcal{P}|
    }.
\end{equation}

This quantity measures the probability that a culturally relevant constraint is maintained at an arbitrary applicable turn.
Because anchors differ in how frequently they are applicable, we additionally compute a macro-averaged AMR. Let $\mathcal{P}_a$ denote the applicable turns for anchor $a$. Then
\begin{equation}
    \mathrm{AMR}_a
    =
    \frac{1}{|\mathcal{P}_a|}
    \sum_{t\in\mathcal{P}_a} M_{a,t},
\end{equation}
and
\begin{equation}
    \mathrm{AMR}_{\mathrm{macro}}
    =
    \frac{1}{|\mathcal{A}|}
    \sum_{a\in\mathcal{A}}
    \mathrm{AMR}_a.
\end{equation}

Micro AMR emphasizes frequently applicable constraints, whereas macro AMR assigns equal weight to each anchor. We report both metrics over all anchors.

\subsection{Contradiction Rate}

To distinguish direct contradiction from other violations, we report the contradiction rate:
\begin{equation}
    R_{\mathrm{CON}}
    =
    \frac{
        \sum_{(a,t)\in\mathcal{P}}
        \mathbb{I}[Y_{a,t}=\texttt{CONTRADICTED}]
    }{
        |\mathcal{P}|
    }.
\end{equation}

\subsection{Temporal Survival Analysis}

Turn-level AMR does not capture when cultural consistency deteriorates. We therefore construct an ordered trajectory for every anchor that is applicable in at least two turns, and denote the set of such anchors by $\mathcal{A}_{\mathrm{traj}} \subseteq \mathcal{A}$. For $a \in \mathcal{A}_{\mathrm{traj}}$, we write $M_{a,r}$ for its maintenance indicator at the $r$-th applicable turn, where $r=1,\ldots,R_a$ and $R_a = |\mathcal{P}_a|$. The first cultural violation time is
\begin{equation}
    T_a = \min\{r : M_{a,r} = 0\}.
\end{equation}

If no violation occurs before the end of the conversation, the trajectory is right-censored at $R_a$. Given $d_r$ first-violation events among $n_r$ anchors still at risk at relevant turn $r$, cultural-consistency survival is estimated using the Kaplan--Meier estimator:
\begin{equation}
    \widehat{S}(r)
    =
    \prod_{j\leq r}
    \left(
        1-\frac{d_j}{n_j}
    \right).
\end{equation}

Here, $\widehat{S}(r)$ denotes the estimated probability that an anchor remains culturally consistent through at least $r$ applicable turns. We report $\widehat{S}(3)$ and $\widehat{S}(5)$. AMR and survival capture complementary properties: AMR measures average maintenance across applicable decisions, whereas survival measures the durability of cultural consistency before the first failure.

\subsection{Recovery and Persistence}

For every anchor that experiences at least one violation, recovery is recorded only when a maintained state occurs after the first violation and the anchor remains maintained at its final applicable turn. Let $\mathcal{E}$ denote the set of anchors with at least one violation. We define
\begin{equation}
    R_{\mathrm{rec}}
    =
    \frac{1}{|\mathcal{E}|}
    \sum_{a\in\mathcal{E}}
    \mathbb{I}\!\left[
    \begin{array}{c}
        \exists r>T_a:\ M_{a,r}=1,\\
        M_{a,R_a}=1
    \end{array}
    \right].
\end{equation}

To quantify violation persistence, we let $L_a$ denote the length of the longest run of consecutive applicable turns in which $M_{a,r}=0$, with $L_a = 0$ if anchor $a$ is never violated.
The reported mean maximum violation-run length is
\begin{equation}
    \overline{L}
    =
    \frac{1}{|\mathcal{A}_{\mathrm{traj}}|}
    \sum_{a\in\mathcal{A}_{\mathrm{traj}}}
    L_a.
\end{equation}

A high recovery rate combined with a short violation run indicates temporary cultural weakening, whereas a low recovery rate together with long violation runs indicates persistent drift.

\section{Training Configuration}
\label{sec:training_con}

Table~\ref{tab:training-config} summarizes the training hyperparameters for SFT and GRPO in our framework. SFT is used to ensure that the model consistently produces the desired output format before GRPO, thereby supporting stable GRPO training, rather than to provide task-specific supervision for dialogue. Both SFT and GRPO data are drawn from short conversations in the CultureConverse training set. All models share the same training hyperparameters.

\begin{table*}[htbp] \centering \small \setlength{\tabcolsep}{12pt} \renewcommand{\arraystretch}{1.15} \begin{tabular}{lll} \toprule \textbf{Configuration} & \textbf{SFT} & \textbf{GRPO} \\ \midrule GPU & H100 & H100 \\ Epochs & 1 & 1 \\ Warmup ratio & 0.05 & 0.05 \\ LoRA rank / alpha & 32 / 64 & 32 / 64 \\ LoRA target modules & \texttt{q/k/v/o\_proj} & \texttt{q/k/v/o\_proj} \\ Initial learning rate & $2 \times 10^{-4}$ & $5 \times 10^{-6}$ \\ Rollouts per prompt & N/A & 12 \\ Rollout temperature & N/A & 1 \\ KL loss coefficient / type & N/A & 0.01 / \texttt{low\_var\_kl} \\ Entropy coefficient & N/A & 0.001 \\ Training data & 6,000 single-turn examples & 9,498 stage-level instances \\ Validation samples & 500 & 500 \\ Reward model & N/A & DeepSeek-V3.2 \\ \bottomrule \end{tabular} \caption{Training configurations for SFT and GRPO. N/A denotes not applicable.} \label{tab:training-config} \end{table*}

\section{Bootstrap confidence intervals.} 
\label{sec:Bootstrap confidence intervals}
We compute 95\% confidence intervals using episode-level percentile bootstrapping. For each backbone, configuration, and language split, we independently draw 10,000 bootstrap samples with replacement, each containing the same number of conversations as the original sample. All turns and cultural-anchor evaluations within each conversation are sampled together to preserve within-conversation dependence. We recompute each metric on every bootstrap sample and use the 2.5th and 97.5th percentiles of the resulting estimates as the confidence interval bounds.

\paragraph{Paired bootstrap confidence intervals for ablation.} The 95\% confidence intervals of +DCM and +SCCA-GRPO in Table~3 overlap for several metrics, leaving uncertainty about the incremental contribution of SCCA-GRPO. To directly quantify these differences, we compute paired cluster bootstrap confidence intervals. We match dialogues from +DCM and CulSTAR by scenario ID and perform 10,000 paired bootstrap resamples using dialogues as clusters. For each resample, we recompute response-quality and cultural anchor-tracking metrics for both systems and calculate their paired differences. We construct 95\% confidence intervals using the 2.5th and 97.5th percentiles of the resulting difference distributions. As shown in Table~\ref{tab:ablation_paired_ci}, the intervals exclude zero for improvements in 3H and reductions in contradiction rates in both English and native-language settings. English AMR and recovery also show positive intervals, while the remaining metrics have intervals that include zero. These results provide statistical support for the incremental benefits of SCCA-GRPO, particularly in response quality and contradiction reduction.

\begin{table*}[t]
\centering
\small
\setlength{\tabcolsep}{5pt}
\renewcommand{\arraystretch}{1.15}

\begin{tabular}{@{}lcc@{}}
\toprule
\rowcolor{gray!12}
\textbf{Metric} &
\textbf{English} &
\textbf{Native} \\
\midrule

\multicolumn{3}{@{}l}{\textbf{Response quality}} \\
3H Avg $\uparrow$
& $+0.11\;[+0.05,+0.17]$
& $+0.10\;[+0.04,+0.17]$ \\

TDQ Avg $\uparrow$
& $0.00\;[-0.03,+0.04]$
& $+0.01\;[-0.02,+0.04]$ \\

\midrule
\multicolumn{3}{@{}l}{\textbf{Cultural anchor tracking}} \\

AMR $\uparrow$
& $+2.2\;[+0.2,+4.1]$
& $+1.6\;[-0.9,+4.1]$ \\

$\widehat{S}(5)$ $\uparrow$
& $+3.0\;[-0.1,+6.2]$
& $+2.2\;[-1.1,+5.4]$ \\

Contra. $\downarrow$
& $-4.0\;[-5.3,-2.7]$
& $-4.5\;[-5.77,-3.27]$ \\

Recovery $\uparrow$
& $+4.8\;[+1.0,+8.7]$
& $+2.6\;[-1.3,+6.4]$ \\

\bottomrule
\end{tabular}

\caption{
Paired cluster bootstrap analysis of the incremental
effect of SCCA-GRPO over DCM.
Values report paired differences between CulSTAR
and +DCM with 95\% confidence intervals from
10,000 bootstrap resamples matched by scenario ID.
Response-quality differences are in score points,
while cultural anchor-tracking differences are
in percentage points.
}

\label{tab:ablation_paired_ci}
\end{table*}

\section{Tool Construction.}
Our framework integrates four specialized tools
for cultural knowledge retrieval, taboo identification,
and multilingual assistance, as summarized in
Table~\ref{tab:tool_construction}.
Cultural Knowledge Retrieval uses Perplexity web
search tool to return the five most relevant results
for each model-generated query, ensuring that
retrieval is guided by the agent's information
needs. Taboo Retrieval accesses a curated database
of 1,268 taboo-related entries filtered from
NormAd~\citep{rao2025normad}, EtiCor++~\citep{dwivedi2025eticor++}, and CultureBank~\citep{shi2024culturebank}.
For multilingual assistance, we employ
Gemma-SEA-LION-v3-9B-IT with two task-specific
prompts. The Multilingual Expert Model provides
faithful English paraphrases of non-English or
code-mixed inputs and identifies culturally
sensitive terms, while Low-resource Language
Polishing improves response fluency in the
original language without altering its meaning
or content. The complete prompts are provided in
Table~\ref{tab:language_tool_prompts}.

\begin{table*}[t]
\centering
\small
\setlength{\tabcolsep}{5pt}
\renewcommand{\arraystretch}{1.12}
\begin{tabular}{@{}p{0.22\textwidth}p{0.30\textwidth}p{0.42\textwidth}@{}}
\toprule
\textbf{Tool} & \textbf{Implementation} & \textbf{Function and Output} \\
\midrule

Cultural Knowledge Retrieval
& Perplexity web search tool
& Retrieves the top five results most relevant to
model-generated queries, providing cultural
knowledge without directly querying raw user input. \\

\addlinespace[3pt]

Taboo Retrieval
& Curated database of 1,268 taboo-related entries
from NormAd, EtiCor++, and CultureBank
& Retrieves relevant cultural taboos to support
context-sensitive decisions and avoid
culturally inappropriate responses. \\

\addlinespace[3pt]

Multilingual Expert Model
& Gemma-SEA-LION-v3-9B-IT with a
language-comprehension prompt
& Produces faithful English paraphrases of
non-English or code-mixed input and identifies
culturally sensitive terms with brief explanations
and sensitivity levels in structured JSON. \\

\addlinespace[3pt]

Low-resource Language Polishing
& Gemma-SEA-LION-v3-9B-IT with a
language-polishing prompt
& Improves the fluency and clarity of generated
responses while preserving the original language,
meaning, and information. \\

\bottomrule
\end{tabular}
\caption{Implementation and functionality of the four
external tools used in our framework.}
\label{tab:tool_construction}
\end{table*}

\begin{table*}[t]
\centering
\footnotesize
\setlength{\tabcolsep}{4pt}
\renewcommand{\arraystretch}{1.12}

\begin{tabularx}{\textwidth}{
@{}p{0.19\textwidth}
p{0.12\textwidth}
>{\raggedright\arraybackslash}X@{}}
\toprule
\textbf{Tool} &
\textbf{Prompt Type} &
\textbf{Prompt Content} \\
\midrule

\textbf{Low-resource Language Polishing}
& System
& You are a text polisher. You receive a draft
reply and return a cleaner, more fluent version
of it in the SAME language as the draft. You
never translate, never switch languages, never
change the meaning, and never add or remove
information. \\

\midrule

\textbf{Multilingual Expert Model}
& System
& You are UseHelper, a Southeast Asian language
comprehension aid embedded in a larger agent.
Your ONLY job is to make non-English or code-mixed
input understandable to an English-reasoning policy.
You are not the assistant and you do not answer
the user. \\

\addlinespace[3pt]

& User
& \begin{minipage}[t]{\linewidth}
\raggedright
Paraphrase the user input below into clear,
faithful English.

\smallskip
Return ONLY a JSON object, no preamble,
no markdown, no explanation:

\smallskip
\ttfamily
\{
\par\hspace*{1em}"english\_paraphrase":
"\textless one faithful English rendering,
1-3 sentences, no commentary\textgreater",
\par\hspace*{1em}"cultural\_terms": [
\par\hspace*{2em}\{
"term": "\textless original term\textgreater",
\par\hspace*{2em}"explanation":
"\textless\(\leq\)12 words\textgreater",
\par\hspace*{2em}"sensitivity": "low|medium|high"
\}
\par\hspace*{1em}]
\par\}

\smallskip
\normalfont
Rules:
\par - Include a term in cultural\_terms ONLY
if it is culturally, religiously, or socially
loaded. Skip ordinary vocabulary. If there
are none, return an empty list.
\par - Do NOT translate word-by-word,
do NOT add advice, do NOT answer the user.
\par - Keep every field as short as possible.

\smallskip
\texttt{USER INPUT:}
\end{minipage} \\

\bottomrule
\end{tabularx}

\caption{Task-specific prompts for the two
language-assistance tools. Both tools use
Gemma-SEA-LION-v3-9B-IT as the backbone.}
\label{tab:language_tool_prompts}
\end{table*}

\section{Human Annotation and LLM-as-Judge Validation}
\label{app:human evaluation}

\begin{table*}[t]
\centering
\small
\setlength{\tabcolsep}{9pt}
\renewcommand{\arraystretch}{1.1}
\begin{tabular}{@{}l ccc cc@{}}
\toprule
& \multicolumn{3}{c}{\textbf{Language-Macro Cohen's $\kappa$}}
& \multicolumn{2}{c}{\textbf{Pass B Coverage}} \\
\cmidrule(lr){2-4}
\cmidrule(l){5-6}
\textbf{LLM Judge}
& \textbf{Pass A}
& \textbf{Pass B (5-class)}
& \textbf{Pass B (Binary)}
& \textbf{5-class}
& \textbf{Binary} \\
\midrule
DeepSeek-Chat-V3.1
& 0.416 & 0.269 & 0.311 & 165 & 176 \\
GPT-5-mini
& 0.093 & 0.160 & 0.289 & 266 & 279 \\
GPT-5.4
& 0.628 & 0.298 & 0.271 & 259 & 272 \\
\textbf{GPT-5.5}
& \textbf{0.734}
& \textbf{0.516}
& \textbf{0.665}
& \textbf{284}
& \textbf{298} \\
\bottomrule
\end{tabular}

\caption{Comparison of four LLM judges across ten
languages. Cohen's $\kappa$ is macro-averaged across
languages. Pass B coverage counts samples jointly
identified as applicable by the human annotators
and the corresponding judge. The human-applicable
subset contains 307 pairs. Five-class evaluation
additionally excludes unresolved human majority ties.}
\label{tab:judge_selection}
\end{table*}

\begin{table*}[t]
\centering
\small
\setlength{\tabcolsep}{4.5pt}
\renewcommand{\arraystretch}{1.08}

\begin{tabular}{@{}l r cc r c cc cc@{}}
\toprule
&
\multicolumn{3}{c}{\textbf{Pass A: Applicability}}
&
\multicolumn{6}{c}{\textbf{Pass B: Compliance}} \\
\cmidrule(lr){2-4}
\cmidrule(l){5-10}

&
&
&
&
&
&
\multicolumn{2}{c}{\textbf{5-class}}
&
\multicolumn{2}{c}{\textbf{Binary}} \\
\cmidrule(lr){7-8}
\cmidrule(l){9-10}

\textbf{Language}
& $n_A$
& \textbf{Human $\kappa_F$}
& \textbf{GPT $\kappa_C$}
& $n_B^H$
& $n_B^{G,5}/n_B^{G,2}$
& \textbf{Human $\kappa_F$}
& \textbf{GPT $\kappa_C$}
& \textbf{Human $\kappa_F$}
& \textbf{GPT $\kappa_C$} \\
\midrule
English
& 100 & 0.412 & 0.503 & 13 & 11/13
& 0.240 & 0.482 & 0.494 & 0.519 \\

Indonesian
& 166 & 0.510 & 0.744 & 54 & 51/54
& 0.316 & 0.540 & 0.438 & 0.567 \\

Japanese
& 43 & 0.515 & 0.718 & 24 & 24/24
& 0.110 & 0.524 & 0.356 & 0.833 \\

Korean
& 73 & 0.685 & 0.866 & 13 & 11/12
& 0.401 & 0.722 & 0.422 & 0.571 \\

Malay
& 110 & 0.585 & 0.567 & 34 & 30/31
& 0.258 & 0.249 & 0.677 & 0.739 \\

Filipino
& 60 & 0.754 & 0.867 & 27 & 24/25
& 0.345 & 0.308 & 0.264 & 0.627 \\

Tamil
& 68 & 0.597 & 0.547 & 29 & 28/28
& 0.304 & 0.796 & 0.448 & 0.928 \\

Chinese
& 103 & 0.801 & 0.768 & 36 & 34/35
& 0.475 & 0.492 & 0.848 & 0.928 \\

Thai
& 77 & 0.830 & 0.879 & 18 & 16/18
& 0.197 & 0.448 & 0.313 & 0.556 \\

Vietnamese
& 94 & 0.801 & 0.877 & 59 & 55/58
& 0.092 & 0.599 & 0.199 & 0.378 \\

\midrule
\textbf{Pooled}
& \textbf{894}
& \textbf{0.660}
& \textbf{0.737}
& \textbf{307}
& \textbf{284/298}
& \textbf{0.358}
& \textbf{0.570}
& \textbf{0.605}
& \textbf{0.729} \\
\bottomrule
\end{tabular}

\caption{Human inter-annotator agreement and
GPT-5.5 alignment with human annotations across
ten languages. $\kappa_F$ denotes Fleiss' kappa
among three human annotators, while $\kappa_C$
denotes Cohen's kappa between GPT-5.5 and
human majority labels. $n_B^H$ counts pairs
unanimously marked applicable by human annotators.
$n_B^{G,5}$ and $n_B^{G,2}$ denote the corresponding
judge evaluation sizes for five-class and binary
compliance. Pooled kappa values are calculated
over combined samples.}
\label{tab:language_judge_agreement}
\end{table*}

\paragraph{Human annotation.}
To assess the reliability of our evaluation framework, we conducted human annotation across English and nine native languages, covering 894 anchor--turn pairs. Each language was evaluated independently by three annotators proficient in the language and familiar with its cultural context. Human agreement was assessed using Fleiss' $\kappa$ for both anchor applicability and cultural compliance.

\paragraph{LLM judge selection.}
We compared four candidate judges, including DeepSeek-Chat-V3.1, GPT-5-mini, GPT-5.4, and GPT-5.5, against human majority judgments. As shown in Table~\ref{tab:judge_selection}, GPT-5.5 achieved the highest language-macro agreement across all three evaluation settings and led in binary compliance agreement for eight of the ten languages. It also provided the broadest coverage of human-applicable pairs. Table~\ref{tab:language_judge_agreement} further shows strong overall alignment between GPT-5.5 and human judgments, with pooled Cohen's $\kappa$ values of 0.737 for applicability and 0.729 for binary compliance. Fine-grained five-class judgments exhibited lower agreement, reflecting greater annotation ambiguity. Importantly, our primary cultural-drift metrics, including AMR, survival, and recovery, rely on binary maintenance judgments. The five annotation categories are mapped into maintained and violated states, so disagreements within the same binary category do not affect these metrics. This distinction is reflected in the stronger agreement on binary compliance ($\kappa=0.729$) compared with fine-grained classification ($\kappa=0.570$). Direct contradiction rates are evaluated separately using the corresponding fine-grained label. These findings support our selection of GPT-5.5 as the automated judge for evaluating cultural consistency across multilingual conversations.

\paragraph{End-to-end judgment validation.}
We further assessed end-to-end judgment reliability by combining applicability and compliance decisions into three categories: Not Applicable, Compliant, and Violated. Across 871 anchor--turn pairs with resolved human gold labels, GPT-5.5 achieved 81.06\% accuracy and a macro-F1 of 0.783. Cultural violation detection reached 0.849 recall and 0.696 F1, indicating high sensitivity to violations alongside a tendency toward false positives.

\section{Detailed Results}
\label{app:detailed_results}

Tables~\ref{tab:main_results}, ~\ref{tab:long_detailed_add}, and ~\ref{tab:cura_ablation_detailed_add} provide detailed breakdowns and supplementary results corresponding to Tables~\ref{tab:main_results_shorter}, ~\ref{tab:long_results}, and ~\ref{tab:cura_ablation}, respectively.

\begin{table*}[t]
\centering
\scriptsize
\setlength{\tabcolsep}{5pt}
\renewcommand{\arraystretch}{0.9}
\resizebox{\textwidth}{!}{%
\begin{tabular}{ll ccc l ccc l l}
\toprule
\rowcolor{gray!12}
 & & \multicolumn{3}{c}{\textbf{3H components}} & \multicolumn{1}{c}{\cellcolor{hhcol}\textbf{3H}}
 & \multicolumn{3}{c}{\textbf{TDQ components}} & \multicolumn{1}{c}{\cellcolor{tdqcol}\textbf{TDQ}} & \\
\rowcolor{gray!12}
\multirow{-2}{*}{\textbf{Backbone}} & \multirow{-2}{*}{\textbf{Method}}
 & Help & Hon & Harm & \multicolumn{1}{c}{\cellcolor{hhcol}Avg}
 & Nat & Plaus & Typ & \multicolumn{1}{c}{\cellcolor{tdqcol}Avg} & \multicolumn{1}{c}{\multirow{-2}{*}{\textbf{Clean \%}}} \\
\midrule

\multicolumn{11}{c}{\cellcolor{gray!12}\textit{English}} \\
\midrule

\rowcolor{fam2}  & Base & 3.03 & 3.39 & 3.64 & \cellcolor{hhcol}3.35\phantom{\kern1pt \up{+00.0\%}} & 3.88 & 3.01 & \cellcolor{bestcol}\textbf{3.57} & \cellcolor{tdqcol}3.49\phantom{\kern1pt \up{+00.0\%}} & 67.2\phantom{\kern1pt \up{+00.0\%}} \\
\rowcolor{fam2}  & + SFT & \cellcolor{bestcol}\textbf{3.05} & 3.50 & 3.54 & \cellcolor{hhcol}3.36\kern1pt \up{+0.3\%} & 3.88 & 2.99 & 3.49 & \cellcolor{tdqcol}3.45\kern1pt \down{-1.1\%} & 75.8\kern1pt \up{+12.8\%} \\
\rowcolor{fam2} \multirow{-3}{*}{Qwen3-8B-IT} & \textbf{CulSTAR} & 3.03 & \cellcolor{bestcol}\textbf{3.69} & \cellcolor{bestcol}\textbf{3.75} & \cellcolor{bestcol}\textbf{3.49}\kern1pt \up{+4.2\%} & \cellcolor{bestcol}\textbf{3.90} & \cellcolor{bestcol}\textbf{3.11} & 3.53 & \cellcolor{bestcol}\textbf{3.51}\kern1pt \up{+0.6\%} & \cellcolor{bestcol}\textbf{79.2}\kern1pt \up{+17.9\%} \\
\cmidrule(lr){1-11}

\rowcolor{fam3}  & Base & 2.70 & 3.16 & 3.23 & \cellcolor{hhcol}3.03\phantom{\kern1pt \up{+00.0\%}} & 3.72 & 2.76 & \cellcolor{bestcol}\textbf{3.56} & \cellcolor{tdqcol}3.35\phantom{\kern1pt \up{+00.0\%}} & 70.2\phantom{\kern1pt \up{+00.0\%}} \\
\rowcolor{fam3}  & + SFT & \cellcolor{bestcol}\textbf{3.15} & 3.13 & 3.29 & \cellcolor{hhcol}3.19\kern1pt \up{+5.3\%} & \cellcolor{bestcol}\textbf{3.83} & 2.73 & 3.50 & \cellcolor{tdqcol}3.35\kern1pt \same{0.0\%} & 67.4\kern1pt \down{-4.0\%} \\
\rowcolor{fam3} \multirow{-3}{*}{Llama-3.1-8B-IT} & \textbf{CulSTAR} & 2.92 & \cellcolor{bestcol}\textbf{3.54} & \cellcolor{bestcol}\textbf{3.67} & \cellcolor{bestcol}\textbf{3.38}\kern1pt \up{+11.6\%} & 3.77 & \cellcolor{bestcol}\textbf{3.02} & 3.54 & \cellcolor{bestcol}\textbf{3.44}\kern1pt \up{+2.7\%} & \cellcolor{bestcol}\textbf{78.4}\kern1pt \up{+11.7\%} \\
\cmidrule(lr){1-11}

\rowcolor{fam4}  & Base & 2.28 & 3.01 & 3.19 & \cellcolor{hhcol}2.83\phantom{\kern1pt \up{+00.0\%}} & 3.10 & 2.74 & 3.53 & \cellcolor{tdqcol}3.12\phantom{\kern1pt \up{+00.0\%}} & 75.2\phantom{\kern1pt \up{+00.0\%}} \\
\rowcolor{fam4}  & + SFT & 2.26 & 3.01 & 3.18 & \cellcolor{hhcol}2.82\kern1pt \down{-0.4\%} & 3.05 & 2.72 & 3.53 & \cellcolor{tdqcol}3.10\kern1pt \down{-0.6\%} & \cellcolor{bestcol}\textbf{78.4}\kern1pt \up{+4.3\%} \\
\rowcolor{fam4} \multirow{-3}{*}{SEA-LION-v4-8B-IT} & \textbf{CulSTAR} & \cellcolor{bestcol}\textbf{2.84} & \cellcolor{bestcol}\textbf{3.35} & \cellcolor{bestcol}\textbf{3.62} & \cellcolor{bestcol}\textbf{3.27}\kern1pt \up{+15.5\%} & \cellcolor{bestcol}\textbf{3.72} & \cellcolor{bestcol}\textbf{2.93} & \cellcolor{bestcol}\textbf{3.55} & \cellcolor{bestcol}\textbf{3.40}\kern1pt \up{+9.0\%} & 77.6\kern1pt \up{+3.2\%} \\

\midrule
\multicolumn{11}{c}{\cellcolor{gray!12}\textit{Native}} \\
\midrule

\rowcolor{fam2}  & Base & 2.81 & 3.33 & 3.56 & \cellcolor{hhcol}3.23\phantom{\kern1pt \up{+00.0\%}} & 3.75 & 2.91 & \cellcolor{bestcol}\textbf{3.58} & \cellcolor{tdqcol}3.41\phantom{\kern1pt \up{+00.0\%}} & 77.0\phantom{\kern1pt \up{+00.0\%}} \\
\rowcolor{fam2}  & + SFT & 2.99 & 3.57 & 3.59 & \cellcolor{hhcol}3.38\kern1pt \up{+4.6\%} & \cellcolor{bestcol}\textbf{3.77} & 2.94 & 3.54 & \cellcolor{tdqcol}3.42\kern1pt \up{+0.3\%} & 77.4\kern1pt \up{+0.5\%} \\
\rowcolor{fam2} \multirow{-3}{*}{Qwen3-8B-IT} & \textbf{CulSTAR} & \cellcolor{bestcol}\textbf{3.00} & \cellcolor{bestcol}\textbf{3.67} & \cellcolor{bestcol}\textbf{3.74} & \cellcolor{bestcol}\textbf{3.47}\kern1pt \up{+7.4\%} & 3.75 & \cellcolor{bestcol}\textbf{3.03} & 3.57 & \cellcolor{bestcol}\textbf{3.45}\kern1pt \up{+1.2\%} & \cellcolor{bestcol}\textbf{77.8}\kern1pt \up{+1.0\%} \\
\cmidrule(lr){1-11}

\rowcolor{fam3}  & Base & 2.17 & 2.42 & 2.78 & \cellcolor{hhcol}2.46\phantom{\kern1pt \up{+00.0\%}} & 2.99 & 2.42 & 3.53 & \cellcolor{tdqcol}2.98\phantom{\kern1pt \up{+00.0\%}} & 63.0\phantom{\kern1pt \up{+00.0\%}} \\
\rowcolor{fam3}  & + SFT & 2.62 & 2.62 & 2.95 & \cellcolor{hhcol}2.73\kern1pt \up{+11.0\%} & 3.01 & 2.46 & 3.52 & \cellcolor{tdqcol}3.00\kern1pt \up{+0.7\%} & 61.6\kern1pt \down{-2.2\%} \\
\rowcolor{fam3} \multirow{-3}{*}{Llama-3.1-8B-IT} & \textbf{CulSTAR} & \cellcolor{bestcol}\textbf{2.85} & \cellcolor{bestcol}\textbf{3.41} & \cellcolor{bestcol}\textbf{3.47} & \cellcolor{bestcol}\textbf{3.24}\kern1pt \up{+31.7\%} & \cellcolor{bestcol}\textbf{3.56} & \cellcolor{bestcol}\textbf{2.86} & \cellcolor{bestcol}\textbf{3.59} & \cellcolor{bestcol}\textbf{3.34}\kern1pt \up{+12.1\%} & \cellcolor{bestcol}\textbf{73.0}\kern1pt \up{+15.9\%} \\
\cmidrule(lr){1-11}

\rowcolor{fam4}  & Base & 2.20 & 2.95 & 3.14 & \cellcolor{hhcol}2.76\phantom{\kern1pt \up{+00.0\%}} & 2.97 & 2.64 & \cellcolor{bestcol}\textbf{3.57} & \cellcolor{tdqcol}3.06\phantom{\kern1pt \up{+00.0\%}} & 74.2\phantom{\kern1pt \up{+00.0\%}} \\
\rowcolor{fam4}  & + SFT & 2.22 & 2.98 & 3.19 & \cellcolor{hhcol}2.80\kern1pt \up{+1.4\%} & 2.94 & 2.65 & 3.54 & \cellcolor{tdqcol}3.04\kern1pt \down{-0.7\%} & 73.8\kern1pt \down{-0.5\%} \\
\rowcolor{fam4} \multirow{-3}{*}{SEA-LION-v4-8B-IT} & \textbf{CulSTAR} & \cellcolor{bestcol}\textbf{2.82} & \cellcolor{bestcol}\textbf{3.43} & \cellcolor{bestcol}\textbf{3.62} & \cellcolor{bestcol}\textbf{3.29}\kern1pt \up{+19.2\%} & \cellcolor{bestcol}\textbf{3.59} & \cellcolor{bestcol}\textbf{2.90} & 3.56 & \cellcolor{bestcol}\textbf{3.35}\kern1pt \up{+9.5\%} & \cellcolor{bestcol}\textbf{77.2}\kern1pt \up{+4.0\%} \\

\bottomrule
\end{tabular}}
\caption{Results on the CultureConverse test set~\citep{tan2026cultureconverse}. Clean \% is the share of clean episodes out of 500. Small numbers next to 3H Avg, TDQ Avg and Clean \% give the relative change over the base model of the same backbone. The best value within each backbone is in bold and highlighted.}
\label{tab:main_results}
\end{table*}

\begin{table*}[t]
\centering
\scriptsize
\setlength{\tabcolsep}{4.5pt}
\renewcommand{\arraystretch}{0.9}
\resizebox{\textwidth}{!}{%
\begin{tabular}{ll ccc ccc ccc}
\toprule

\rowcolor{gray!12}
& &
\multicolumn{3}{c}{\textbf{3H}}
& \multicolumn{3}{c}{\textbf{TDQ}}
& \multicolumn{3}{c}{\textbf{Cultural anchor tracking}} \\

\rowcolor{gray!12}
\multirow{-2}{*}{\textbf{Backbone}}
& \multirow{-2}{*}{\textbf{Method}}
& \multicolumn{1}{c}{\textbf{Help.} $\uparrow$}
& \multicolumn{1}{c}{\textbf{Hon.} $\uparrow$}
& \multicolumn{1}{c}{\textbf{Harm.} $\uparrow$}
& \multicolumn{1}{c}{\textbf{Nat.} $\uparrow$}
& \multicolumn{1}{c}{\textbf{Plaus.} $\uparrow$}
& \multicolumn{1}{c}{\textbf{Typ.} $\uparrow$}
& \multicolumn{1}{c}{\textbf{AMR-M} $\uparrow$}
& \multicolumn{1}{c}{$\widehat{S}(3)$ $\uparrow$}
& \multicolumn{1}{c}{\textbf{MaxRun} $\downarrow$} \\

\midrule
\multicolumn{11}{c}{\cellcolor{gray!12}\textit{English}} \\
\midrule

\rowcolor{fam2}
&
Base
& 2.88
& 2.38
& 3.18
& 2.95
& 2.41
& 3.44
& 66.3
& 41.7
& 2.29 \\

\rowcolor{fam2}
\multirow{-2}{*}{Qwen3-8B-IT}
&
\textbf{Ours}
& \textbf{3.02}
& \textbf{3.39}
& \textbf{3.81}
& \textbf{3.84}
& \textbf{2.91}
& \textbf{3.60}
& \textbf{68.6}
& \textbf{49.6}
& \textbf{1.85} \\

\cmidrule(lr){1-11}

\rowcolor{fam3}
&
Base
& 2.25
& 2.39
& 2.94
& 3.14
& 2.40
& 3.47
& 56.1
& 29.7
& 2.46 \\

\rowcolor{fam3}
\multirow{-2}{*}{Llama-3.1-8B-IT}
&
\textbf{Ours}
& \textbf{2.87}
& \textbf{3.14}
& \textbf{3.68}
& \textbf{3.51}
& \textbf{2.82}
& \textbf{3.60}
& \textbf{74.7}
& \textbf{67.3}
& \textbf{1.41} \\

\cmidrule(lr){1-11}

\rowcolor{fam4}
&
Base
& 2.20
& 2.63
& 3.19
& 2.94
& 2.56
& 3.54
& 62.9
& 34.9
& 2.07 \\

\rowcolor{fam4}
\multirow{-2}{*}{SEA-LION-v4-8B-IT}
&
\textbf{Ours}
& \textbf{2.81}
& \textbf{3.02}
& \textbf{3.55}
& \textbf{3.53}
& \textbf{2.72}
& \textbf{3.59}
& \textbf{70.8}
& \textbf{44.1}
& \textbf{1.99} \\

\midrule
\multicolumn{11}{c}{\cellcolor{gray!12}\textit{Native}} \\
\midrule

\rowcolor{fam2}
&
Base
& 2.70
& 2.50
& 3.21
& 2.90
& 2.44
& 3.54
& 65.0
& 35.2
& 2.35 \\

\rowcolor{fam2}
\multirow{-2}{*}{Qwen3-8B-IT}
&
\textbf{Ours}
& \textbf{2.84}
& \textbf{3.39}
& \textbf{3.78}
& \textbf{3.61}
& \textbf{2.81}
& \textbf{3.61}
& \textbf{79.0}
& \textbf{61.6}
& \textbf{1.42} \\

\cmidrule(lr){1-11}

\rowcolor{fam3}
&
Base
& 2.06
& 2.53
& 3.02
& 2.96
& 2.46
& 3.58
& 47.4
& 20.9
& 2.68 \\

\rowcolor{fam3}
\multirow{-2}{*}{Llama-3.1-8B-IT}
&
\textbf{Ours}
& \textbf{2.55}
& \textbf{3.06}
& \textbf{3.49}
& \textbf{3.20}
& \textbf{2.66}
& \textbf{3.64}
& \textbf{74.4}
& \textbf{51.8}
& \textbf{1.23} \\

\cmidrule(lr){1-11}

\rowcolor{fam4}
&
Base
& 2.15
& 2.79
& 3.23
& 2.84
& 2.55
& \textbf{3.65}
& 61.7
& 34.4
& 1.95 \\

\rowcolor{fam4}
\multirow{-2}{*}{SEA-LION-v4-8B-IT}
&
\textbf{Ours}
& \textbf{2.64}
& \textbf{3.16}
& \textbf{3.61}
& \textbf{3.34}
& \textbf{2.71}
& \textbf{3.65}
& \textbf{71.1}
& \textbf{48.5}
& \textbf{1.88} \\

\bottomrule
\end{tabular}}

\caption{
Detailed results on the \textit{CultureConverse-Long} test set.
3H comprises Helpfulness, Honesty, and Harmlessness, while TDQ comprises
Naturalness, Scenario Plausibility, and Cultural Typicality.
AMR-M denotes macro-averaged anchor maintenance rate, and MaxRun denotes
the mean maximum violation-run length.
Best values within each backbone are bolded.
}
\label{tab:long_detailed_add}
\end{table*}

\definecolor{abbase}{RGB}{246,246,246}
\definecolor{abmem}{RGB}{235,245,250}
\definecolor{abrl}{RGB}{239,236,250}

\begin{table*}[t]
\centering
\scriptsize
\setlength{\tabcolsep}{3.5pt}
\renewcommand{\arraystretch}{0.9}
\resizebox{\textwidth}{!}{%
\begin{tabular}{ll ccc ccc ccc}
\toprule

\rowcolor{gray!12}
& &
\multicolumn{3}{c}{\textbf{3H}}
& \multicolumn{3}{c}{\textbf{TDQ}}
& \multicolumn{3}{c}{\textbf{Cultural anchor tracking}} \\

\rowcolor{gray!12}
\multirow{-2}{*}{\textbf{Backbone}}
& \multirow{-2}{*}{\textbf{Configuration}}
& \multicolumn{1}{c}{\textbf{Help.} $\uparrow$}
& \multicolumn{1}{c}{\textbf{Hon.} $\uparrow$}
& \multicolumn{1}{c}{\textbf{Harm.} $\uparrow$}
& \multicolumn{1}{c}{\textbf{Nat.} $\uparrow$}
& \multicolumn{1}{c}{\textbf{Plaus.} $\uparrow$}
& \multicolumn{1}{c}{\textbf{Typ.} $\uparrow$}
& \multicolumn{1}{c}{\textbf{AMR-M} $\uparrow$}
& \multicolumn{1}{c}{$\widehat{S}(3)$ $\uparrow$}
& \multicolumn{1}{c}{\textbf{MaxRun} $\downarrow$} \\

\midrule
\multicolumn{11}{c}{\cellcolor{gray!12}\textit{English}} \\
\midrule

\rowcolor{abbase}
&
Base
& 2.25
& 2.39
& 2.94
& 3.14
& 2.40
& 3.47
& 56.07
& 29.69
& 2.4626 \\

\rowcolor{abmem}
&
+ DCM
& 2.74
& 2.90
& \textbf{3.72}
& \textbf{3.61}
& 2.76
& 3.56
& 71.81
& 64.05
& 1.5801 \\

\rowcolor{abrl}
\multirow{-3}{*}{Llama-3.1-8B-IT}
&
+ SCCA-GRPO
& \textbf{2.87}
& \textbf{3.14}
& 3.68
& 3.51
& \textbf{2.82}
& \textbf{3.60}
& \textbf{74.66}
& \textbf{67.34}
& \textbf{1.4093} \\

\midrule
\multicolumn{11}{c}{\cellcolor{gray!12}\textit{Native}} \\
\midrule

\rowcolor{abbase}
&
Base
& 2.06
& 2.53
& 3.02
& 2.96
& 2.46
& 3.58
& 47.41
& 20.90
& 2.6778 \\

\rowcolor{abmem}
&
+ DCM
& 2.52
& 2.80
& 3.48
& \textbf{3.29}
& 2.62
& 3.57
& 73.36
& 50.68
& 1.4277 \\

\rowcolor{abrl}
\multirow{-3}{*}{Llama-3.1-8B-IT}
&
+ SCCA-GRPO
& \textbf{2.55}
& \textbf{3.06}
& \textbf{3.49}
& 3.20
& \textbf{2.66}
& \textbf{3.64}
& \textbf{74.44}
& \textbf{51.79}
& \textbf{1.2250} \\

\bottomrule
\end{tabular}}

\caption{
Detailed ablation of DCM and SCCA-GRPO on \textit{CultureConverse-Long} test set.
3H comprises Helpfulness, Honesty, and Harmlessness, while TDQ comprises Naturalness,
Scenario Plausibility, and Cultural Typicality.
AMR-M denotes the macro-averaged anchor maintenance rate,
and MaxRun denotes the mean maximum violation-run length.
Best values within each language setting are bolded.
}
\label{tab:cura_ablation_detailed_add}
\end{table*}

\end{document}